\documentclass[journal]{IEEEtran}
\usepackage{amsmath,url, amssymb}

\usepackage{mathrsfs}
\usepackage[pdftex]{graphicx}
\usepackage[caption=false]{subfig}
\usepackage{balance}
\usepackage{comment}
\usepackage{tabularx}
\usepackage{diagbox}
\usepackage{algorithmic}
\usepackage{array}
\usepackage{enumitem, kantlipsum}
\usepackage[mathscr]{euscript}
 \let\mathscr\relax
\usepackage[scr]{rsfso}
\usepackage{textcomp}
\usepackage{multirow, bm}
\usepackage{makecell}
\usepackage{arydshln}
\usepackage{xcolor}
\usepackage{afterpage}
\usepackage{tikz}
\newcolumntype{?}{!{\vrule width 2\arrayrulewidth}}
\usepackage{amsfonts}
\usepackage{tabularx}
\usepackage{float}
\usepackage{xcolor}
\usepackage{enumitem, kantlipsum}

\def\I{{\mathbf I}}
\def\w{{\mathbf w}}
\def\P{{\mathbf P}}
\def\Ihole{{\bar{\mathbf I}}}
\def\Ihat{{\hat{\mathbf I}}}

\def\D{{{\mathcal{D}}}}

\newcommand{\red}[1]{\textcolor{black}{#1}}
\begin{document}

\title{Interpolation and Denoising of Seismic Data \\ using Convolutional Neural Networks}

\author{Sara~Mandelli,
        Vincenzo~Lipari, 
        Paolo~Bestagini,~\IEEEmembership{Member,~IEEE,}
        and~Stefano~Tubaro,~\IEEEmembership{Senior~Member,~IEEE}
\thanks{S. Mandelli, V. Lipari, P. Bestagini and S. Tubaro are with the Dipartimento di Elettronica, Informazione e Bioingegneria, Politecnico di Milano, Milano 20133, Italy (e-mail: name.surname@polimi.it).}
}

\maketitle

\begin{abstract}
Seismic data processing algorithms greatly benefit from regularly sampled and reliable data. Therefore, interpolation and denoising play a fundamental role as \red{one of} the starting steps of most \red{seismic processing workflows}. We exploit convolutional neural networks for the joint tasks of interpolation and random noise attenuation of 2D common shot gathers. 
Inspired by the great contributions achieved in image processing and computer vision, we investigate a particular architecture of convolutional neural network referred to as \emph{U-net}, which implements a convolutional autoencoder able to describe the complex features of clean and regularly sampled data for reconstructing the corrupted ones. 
In training phase we exploit part of the data for tailoring the network to the specific tasks of interpolation, denoising and joint denoising/interpolation, while during the system deployment we are able to recover the remaining corrupted shot gathers in a computationally efficient procedure. We consider a plurality of data corruptions in our \red{numerical experiments}, including different noise models and different distributions of missing traces. Several examples on synthetic and field data illustrate the \red{appealing features} of the aforementioned strategy. Comparative examples show improvements with respect to \red{recently proposed} solutions for joint denoising and interpolation.
\end{abstract}

\section{Introduction}
%
%


Seismic processing and imaging methods are essential to discover, localize and characterize economically worthwhile geological reservoirs, such as hydrocarbons accumulations, and to manage the extraction of the resources stored in them. 

However, since easy-to exploit resources are dramatically reducing and exploration targets are more and more complex, the requirements for the quality of seismic data, both in term of Signal-to-Noise ratio \red{($\mathrm{S/N}$)} and of regularity and density of its sampling, are constantly increasing.

Unfortunately, \red{various types of random and coherent noise, depending on the environment and on the acquisition technology, often corrupt seismic data sets.}
An additional problem is that economic limitations, cable feathering \red{in marine case}, environmental constraints and elimination of badly acquired traces cause irregular spatial sampling in almost all seismic acquisitions.

Most state-of-the-art seismic processing algorithms, such as reverse-time-migration \cite{chang19903d}, full-waveform-inversion \cite{virieux2009overview} and surface related multiple elimination \cite{ver92} benefit from high quality regularly sampled data.
Consequently, the vast majority of seismic processing \red{workflows} require data pre-processing steps, including effective denoising and trace interpolation algorithms.
Moreover, due to the increasing size of the acquired data, a key factor of these procedures for industrial application is their computational burden, in terms of both memory requirements and computational time.

The problems of trace interpolation and noise attenuation have been widely investigated, either simultaneously or separately.
Among the dozens of interpolation methods have been proposed so far, we can roughly identify four main categories. 

Model-based algorithms implement an implicit migration-demigration pair \cite{stolt2002seismic, fomel2003seismic}.
A major drawback of these techniques is that their performance is strongly affected in case of complex structural burden. 

A second approach, \red{for both denoising and interpolation}, is based on prediction filters \red{\cite{spitz1991seismic,gulunay1986fxdecon,abma1995lateral,liu2012random}}, which assume seismic data to be a (local) superposition of plane w1aves. 
However, these methods target regularly sampled data, which is a heavy limitation.

Due to their repetitive features, clean seismic data are intrinsically low-rank in the time-space domain.
Conversely, noise and missing traces increase the rank of the data \cite{trickett2003f}.
Therefore, algorithms recasting the interpolation (and denoising) problems as rank reduction and matrix/tensor completion have been largely studied in the past decade as third alternative to the problem \cite{trickett2009prestack, oropeza2011simultaneous, yang2013seismic, kumar2014svd, adamo2015irregular}.

A great amount of denoising and interpolation algorithms exploit a transform domain where the clean signal can be represented only by few non-zero coefficients and therefore clean data and noise are more easily separable.
The rationale behind this forth family of methods is that noise and missing traces map in non-sparse artifacts in the transform domain.

Several fixed-basis sparsity-promoting transforms have been widely used also for seismic data interpolation.
Among the various approaches, coming from different fields, we can cite: the Fourier transform \cite{naghizadeh2012seismic,chen2015seismic}, the Hilbert-Huang transform \cite{battista2007application, bekara2009random}, the time-frequency peak filtering, \cite{wu2011noise, liu2013spatiotemporal, tian2014parabolic}, the Radon transform \cite{wang2010seismic}, different curvelet-like transforms \cite{herrmann2008non, gan2015dealiased, wang2014dreamlet, zhang2003physical, herrmann2008sparsity, fomel2010seislet} and the EMD-seislet transform \cite{chen2015emd}.

\red{These methods assume the ability to describe the data in terms of a linear combination of atoms (i.e., the elementary signals) taken from a dictionary (i.e., a predefined set of atoms)}. \red{The aforementioned} transform methods implicitly assume regularity of the data described by analytic models, resulting in \red{predefined} fixed dictionaries. However, these dictionaries can be thought as defining only a subset of the transforms methods.

Alternatively, data driven sparse dictionaries can be learned directly from the dataset.
In other words, these methods assume that clean signals are a linear combination \red{under a sparsity constraint} of the atoms in a learned overcomplete dictionary \red{(i.e., a frame made by a set of $n$ functions with $n > m$, being $m$ the signal dimensions)}.
Learned dictionaries, in the form of explicit matrices for small patches, usually better match the complex data characteristics.
For instance, denoising results obtained using double sparse dictionary learning and outperforming fixed dictionary transforms have been reported in \cite{zhu2015seismic} and \cite{chen2016double}, combining the dictionary learning based sparse transform with the fixed-basis transform, which is called double-sparsity dictionary.
Recently, \cite{zhu2017joint} introduced a joint seismic data denoising and interpolation \red{method} using a masking strategy in the sparse representation of the dictionary.

In \red{recent} years, outstanding advancements brought by deep learning and Convolutional Neural Networks (CNNs) have greatly impacted whole signal and image processing community.
In this context, innovative strategies for data interpolation \red{and} denoising based on deep learning have been proposed in \red{many} image processing tasks.
Indeed, solutions based on CNNs are nowadays often exceeding state-of-the-art results.
However, these methods have barely started to be explored by the geophysical community for the problems of denoising and interpolation.
Promising results for the aforementioned tasks have been reported through residual neural networks \cite{wang2018seismic, jin2018seismic}, generative adversarial networks \cite{siahkoohi2018seismic} and convolutional autoencoders \cite{ mikhailiuk2018deep,mandelli2018seismic}.

In this paper we \red{leverage convolutional autoencoders both to interpolate and to denoise} irregular seismic data  in the shot-gather domain.
\textcolor{black}{In particular, }
inspired by the important contributions achieved in image processing problems, we 
exploit a properly trained \emph{U-net} \cite{ronneberger2015u} as a strongly competitive strategy for noise attenuation and reconstruction of missing traces in pre-stack seismic gathers.
We provide examples on synthetic and field data showing promising performances on either denoising, interpolation, or joint denoising and interpolation problems.
The results obtained for the joint interpolation and denoising task also outperform a recent technique taken as \red{a} reference.


In the following section section we introduce the proposed network architecture \red{and the rationale behind} the use of CNNs for interpolation and denoising of 2D seismic data.
\textcolor{black}{This is followed by}
the specific \red{workflow} to be 
observed
in order to train and test the \emph{U-net} for reconstructing corrupted gathers.
Then,
the proposed method is applied to synthetic and field 2D seismic data. 
\textcolor{black}{Finally,} 
we discuss the advantages and the potential issues of our method and draw conclusions.

\red{We are aware that each research field has its specific jargon. Since the development of deep learning techniques applied to seismic data processing is still in its infancy, the terms used to define the concepts related to CNNs might be not familiar to all the geophysical community. For this reason, we attach as appendix a short glossary in order to clarify some CNN-based terminology which may be unusual for a geophysicist reader.}
\section{Problem Statement and Background on Autoencoders}
\label{sec:intro_method}

In this section, we first report details about the formulation of the tackled problems, namely interpolation and/or denoising of seismic data.
Then, we provide some background concepts on Convolutional Autoencoders (CAs) and how to exploit them for our specific goals, which is useful to understand the rest of the paper.

\subsection{Problem formulation}
\label{subsec:problem}

In this paper, we focus on the problem of reconstructing seismic gathers which have been corrupted by irregular trace sampling and/or additional noise. 
Formally, we represent each original non-corrupted seismic gather as $\I$ \red{and} its corrupted version is denoted as $\Ihole$. 
Our goal is \red{to estimate} a clean and dense version of the seismic data, namely $\Ihat$, as similar as possible to the original corresponding gather $\I$. 

In order to solve this problem, we make use of a particular kind of convolutional neural network named Convolutional Autoencoder (CA).
Our choice is motivated by the great capability of CA in learning compact representations of the data, and by the strong computational efficiency in reconstructing the corrupted ones.  
In the following, we report some backgrounds on CAs, introducing the specific network architecture exploited for the prescribed task.

\subsection{Convolutional Autoencoders for interpolation and denoising}
\label{subsec:unet_intro}

\begin{figure}[t]
	\centering
	\includegraphics[width=0.6\columnwidth]{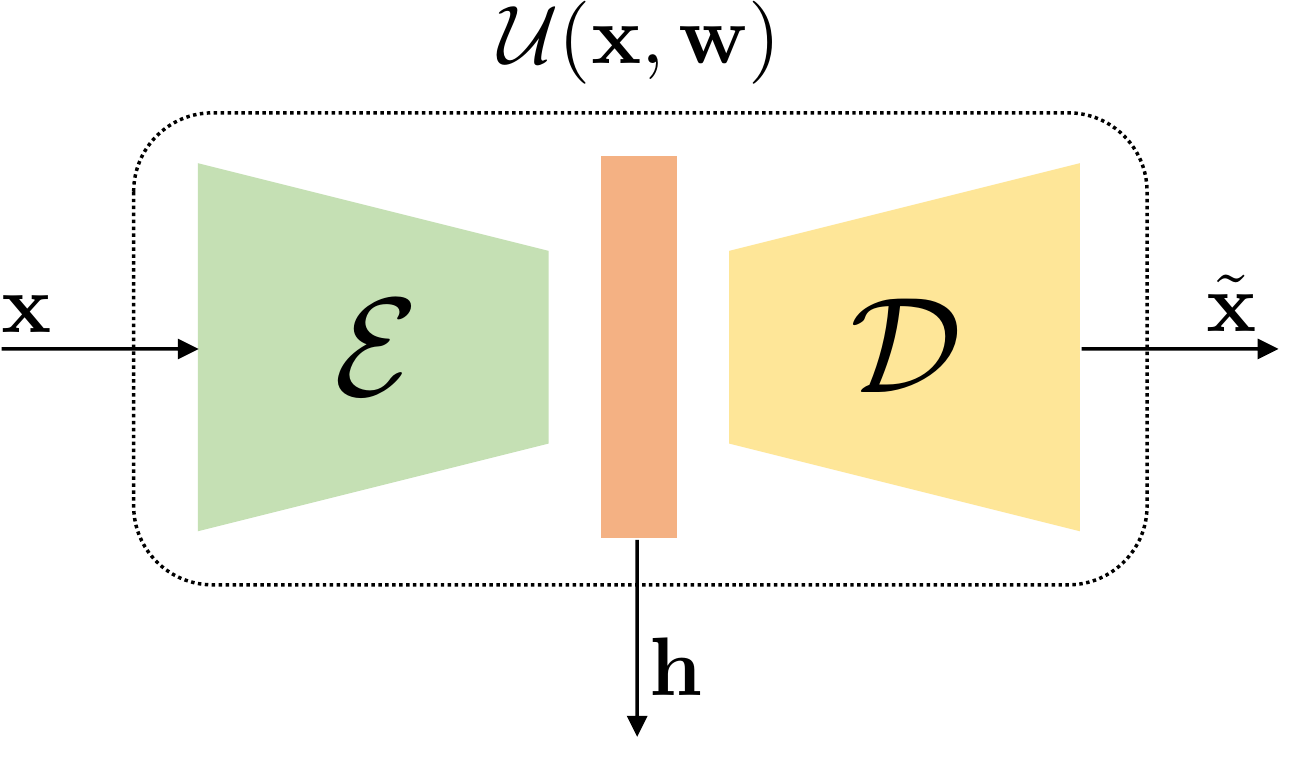}
	\caption{Scheme of a Convolutional Autoencoder architecture. 
	}
	\label{fig:autoencoder}
\end{figure}

Convolutional Autoencoders (CAs) are convolutional neural networks whose architecture can be logically split \red{into} two separate components: the encoder and the decoder.
The CA structure is sketched in \red{Fig.}~\ref{fig:autoencoder}:
\begin{itemize}
	\item the encoder, represented by the operator $\mathcal{E}$, maps the input $\mathbf{x}$ into the \red{so called hidden (or latent)} representation $\mathbf{h}=\mathcal{E}\left(\mathbf{x}\right)$, \red{which is
	the innermost encoding layer of the autoencoder, compressing the input $\mathbf{x}$ into a high-level representation \cite{goodfellow2016};}
	\item the decoder, represented by the operator $\mathcal{D}$, transforms the \red{hidden} representation into an estimate of the input $\tilde{\mathbf{x}}=\mathcal{D}(\mathbf{h})$.
\end{itemize}

For image processing problems, CA proves to be a very powerful instrument for inpainting and denoising tasks \cite{xie2012inpainting, pathak2016context}. 
The rationale behind the use of CA for inpainting and denoising shares some common concepts with the transform-based and dictionary learning techniques. 

\begin{figure*}[t]
\centering
  \includegraphics[width=0.95\textwidth]{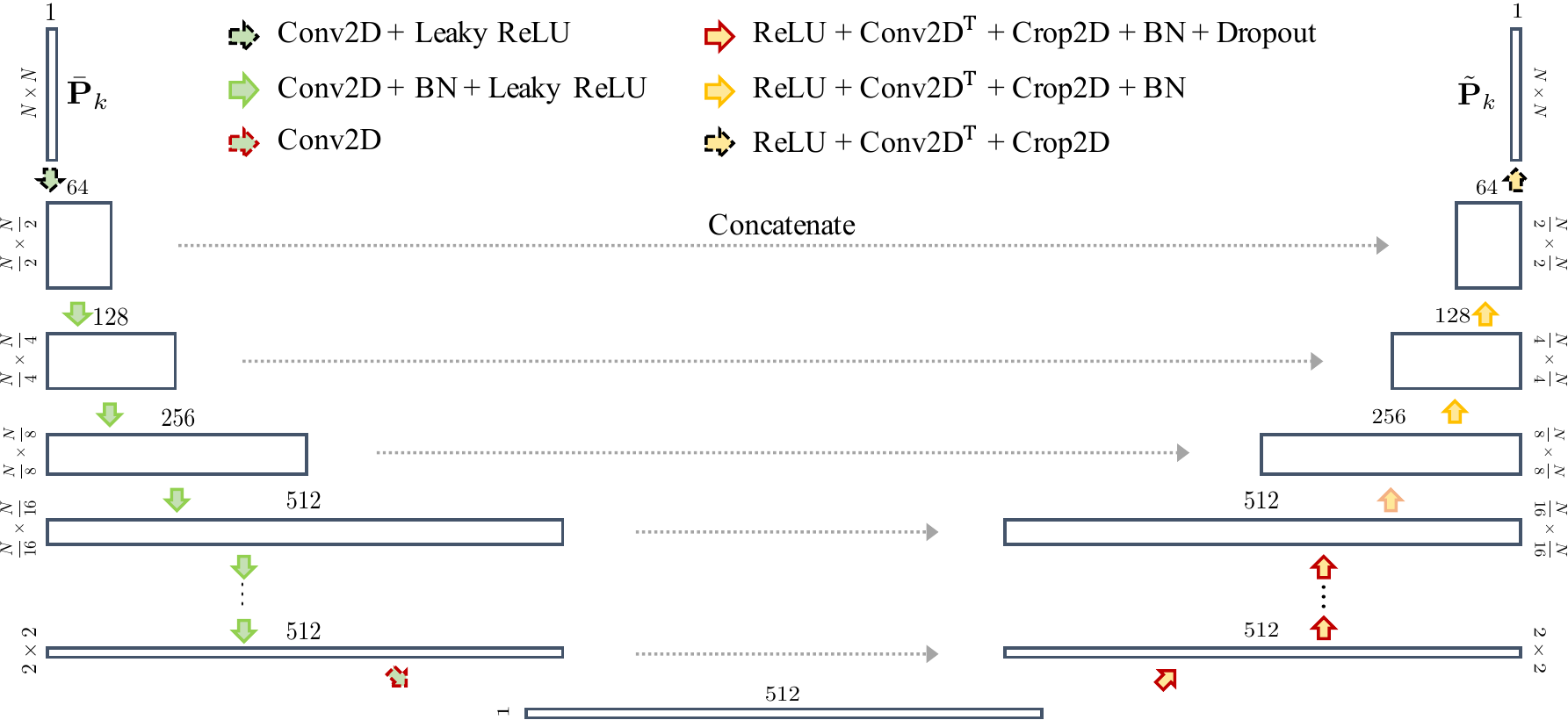}
  \caption{Architecture of the used \emph{U-net}.}
  \label{fig:unet}
\end{figure*}

Indeed, CA is trained so that the encoder part results in a compact representation of clean data, where the interference due to noise and missing samples is not mapped.
Therefore, if the compact representation is correctly built, the result of the decoder is a \emph{dense} clean image without missing samples.
Consequently, it is possible to train a CA to learn a \red{hidden} representation of the clean data in common shot gathers and then to recover clean and densely sampled gathers from noisy and scattered ones.

In particular, in this work we exploit a CNN architecture known as \emph{U-net}.
Originally designed for image segmentation problems and then used for several different tasks \cite{jin2017deep}, the \emph{U-net} is named after the shape it is usually graphically represented with.
Indeed, \emph{U-net} shares a large part of the architecture with classical CA. 
However, in a \emph{U-net}, the representations of the input obtained at different levels of the encoder are directly concatenated to the corresponding decoder levels. \red{This peculiarity typically allows to preserve the structural integrity of the image and to enable a very precise reconstruction.} 
For the sake of brevity, we refer the interested readers to \cite{ronneberger2015u} for a detailed explanation of these architectures.

Continuing the analogy with the transform-based methods, we can think the trained \emph{U-net} as an instrument implicitly providing a multi-scale/multi-resolution \red{compact} representation, able to describe the complex features of clean seismic data where noise and missing data are not modeled.
By using a computer vision terminology, we can think at the interpolation and denoising task as an image transfer problem, with the goal of transforming gathers corrupted by noise and/or missing traces into regularly sampled clean gathers.

\section{Reconstructing the Corrupted Gathers}
\label{sec:algorithm}

In this section we explain how to reconstruct the corrupted seismic gathers.
In particular, we report the technical details concerning the used network architecture and then we describe two \red{strategies} for network training, to be used according to the specific data corruption. 
\red{Finally,} we describe the system deployment, that is \red{the procedure designed to process, after the training stage, the actual gather we want to reconstruct}. 

\subsection{Implementation of U-net}
\label{subsec:unet_implementation}
In order to focus on local portions of the gathers and to ensure a sufficiently large amount of data under analysis, we work in a patch-wise fashion.
Specifically, we divide each \red{shot} gather into $K$ patches of size $N \times N$.
\red{Then, we always consider a single patch as input to the network.
Notice that, for what regards the \emph{U-net} implementation, patches can be extracted from the gather with or without overlap, being the \emph{U-net} architecture independent from the patch selection strategy.}
The CNN we propose takes inspiration by recent works for the interpolation task \cite{mandelli2018seismic} but introduces a few novelties in order to simplify the network architecture at highest levels and enhance the system efficiency without drops in performances. For instance, as seismic data typically has a value range very different from that of natural digital images for which CNNs are typically studied, each patch is \red{multiplied} by a constant gain $G$ \red{(i.e., $ G = 2000$ in the following experiments)}, which proves to be an effective data normalization procedure in terms of \red{convergence speed} and achieved validation results.
\textcolor{black}{Moreover, we do not need to consider the gradient computation and the batch normalization after the first convolutional layer, thus making the proposed approach leaner.}

Considering that corrupted gathers are labeled as $\Ihole$, the generic $k$-th corrupted patch given as input to the network is denoted as $\bar{\P}_k$.


As \emph{U-net} like architectures turn out to be the state-of-the-art for the tasks of image inpainting \cite{YanLLZS18,LiuRSWTC18} and denoising of medical images \cite{heinrich2018residual}, we follow the trend started by \cite{ronneberger2015u}, exploiting a \emph{U-net} architecture composed by the blocks shown in Fig.~\ref{fig:unet}:
\begin{enumerate}
	\item A number of stages where a 2D Convolution with filter size $4\times 4$ and stride $2\times2$, sometimes followed by Batch Normalization (BN) and/or Leaky ReLU, is performed.
	These stages lead to the hidden representation (i.e., the result of the encoding part).
	\red{Specifically, we do not include the BN stage in the outermost layer and after the 2D Convolution leading to $\mathbf{h}$.}
	It is worth noting that the number of filters increases from $64$ to $512$ as we go deep into the network.  
	\item The same number of stages as before where a ReLU, a 2D \red{Transposed} Convolution with filter size $4\times 4$ and stride $2\times2$ and a 2D cropping, possibly followed by BN and Dropout are performed.
	\red{In this case, we use BN on all layers except from the last one, while Dropout is used only in the initial layers, until we reach a patch dimension of $\frac{N}{16} \times \frac{N}{16}$.}
	In each stage we concatenate the result of the corresponding encoding stage as in a typical \emph{U-net} fashion.
	Note that the number of filters is gradually diminished as we go up in the right path of the network (i.e., decoding path).
	The last stage outputs the patch $\tilde{\mathbf{P}}_{k}$, of the same size of the input patch.
\end{enumerate}

Additionally, the overall architecture scales according to the patch dimension $N$, which can be selected depending on possible application-driven constraints.
Anyway, as the network can be characterized by more than $40$ million parameters, it needs to be trained on a significant amount of seismic images as any typical deep learning solutions.

\subsection{U-net training}
\label{subsec:training}

Each corrupted patch $\bar{\mathbf{P}}_{k}$ coming from training dataset $\mathcal{D}_T$ is processed by the \emph{U-net}, and a patch $\tilde{\mathbf{P}}_{k}$ is estimated. The network weights $\mathbf{w}$ are estimated by minimizing the loss function computed with the original uncorrupted patch $\mathbf{P}_{k}$. 
For the task of interpolation only, some samples from $\bar{\mathbf{P}}_{k}$ are already known, therefore we include their knowledge in the loss function.

First, a set of $K$ patches is extracted from image $\Ihole$; each patch $\bar{\mathbf{P}}_{k}$ is processed by the \emph{U-net} architecture $\mathcal{U}(\cdot, \mathbf{w})$, and a patch $\tilde{\mathbf{P}}_{k}$ is estimated. 
For the tasks of denoising only and joint denoising/interpolation, the post-processing block simply divides $\tilde{\mathbf{P}}_{k}$ by the gain $G$. Concerning the task of interpolation only, also the known samples of $\bar{\mathbf{P}}_{k}$ are used together with $\tilde{\mathbf{P}}_{k}$ for reconstructing the patch $\hat{\P}_{k}$. 
Finally, image $\hat{\mathbf{I}}$ is reconstructed by re-assembling together the estimated patches. Whether the patch extraction process required some patch-overlap, the overlapping portions are sample-wise averaged.

Once defined the network architecture, the key point is the design of the training strategy through a proper definition of the cost function tailored to our specific problem.
Indeed, the \emph{U-net} defines a parametric model $ \tilde{\mathbf{x}} = {\mathcal{U}(\mathbf{x}, \mathbf{w})}$, between the output $\tilde{\mathbf{x}}$ and the input $\mathbf{x}$ and network weights $\mathbf{w}$.
The training phase consists in estimating the network weights $ \mathbf{w} $ through the minimization of a distance metrics between the network input and its output.
This distance is usually referred as \emph{loss} function, and its minimization is carried out using iterative techniques (e.g., stochastic gradient methods, etc.).

\begin{figure}[t]
\centering
  \includegraphics[width=0.7\columnwidth]{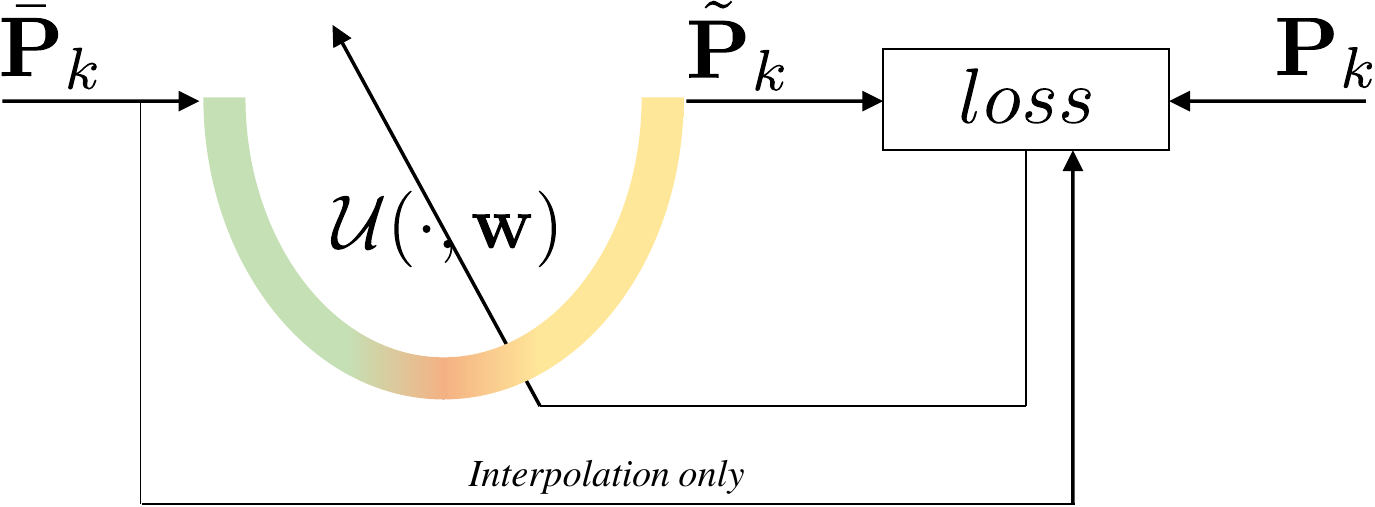}
 \caption{Sketch of the training phase.}
 \label{fig:system_training}
\end{figure}

Specifically, as shown in Fig.~\ref{fig:system_training}, we train the network in order to transform patches $\bar{\P}_k $, extracted from gathers corrupted by noise and/or missing traces, into regularly sampled and clean patches $\tilde{\P}_k$.
As in any supervised learning problem, we assume to have a training dataset $\D_T$ and a validation dataset $\D_V$, each one composed by pairs of corrupted/uncorrupted gathers ($\Ihole, \I)$.

These datasets are exploited for estimating the network parameters $ \mathbf{w} $ and to decide when to stop the cost function minimization process.
Actually, the training stage slightly differs depending of the specific problem: denoising only, interpolation only, and joint denoising/interpolation.

For denoising only and joint denoising/interpolation, model weights are estimated by minimizing the \emph{loss} function between $\mathbf{P}_{k}$ and $\tilde{\mathbf{P}}_{k}$, defined as the squared error over all patches belonging to gathers in the training set $\D_T$.
Formally,
\begin{equation}
	\mathbf{w} = \arg  \min_{\mathbf{w}}  \sum_{\P_k \in \D_T }  \left \| {\P}_k - \tilde{\P}_k (\mathbf{w}) \right \|_\mathrm{F}^2 ,
	\label{eq:training}
\end{equation}
where $ \left \|   \cdot  \right \|_\mathrm{F} $ represents the Frobenius norm and, with a slight abuse of notation, $ \P_k \in \D_T $ denotes patches extracted from gathers belonging to training dataset. 

However, for the task of interpolation only, we a-priori know that only some samples need to be reconstructed by the network (i.e., the missing ones), whereas the others can be left untouched.
For this reason, the \emph{loss} is evaluated only on \red{reconstructed} samples.
This is implicitly performed by adding a masking stage that sets to zero all uncorrupted traces.
Network weights are then estimated as
\begin{equation}
	\mathbf{w} = \arg  \min_{\mathbf{w}}  \sum_{\P_k \in \D_T }  \left \| ({\P}_k - \tilde{\P}_k (\mathbf{w})) \otimes \mathbf{M} \right \|_\mathrm{F}^2 ,
	\label{eq:training_inp}
\end{equation}
where $\otimes$ represents the Hadamard product and $\mathbf{M}$ is a binary mask of size $N \times N$ defined as
\begin{equation}
\left[\mathbf{M}\right]_{i,j}=\begin{cases}
	1& \text{if} \left[\bar{\mathbf{P}}_{k}\right]_{i,j} \text{is a missing sample}\\
	0& \text{otherwise}.
	\end{cases}
	\label{eq:mask}
\end{equation}

As in standard neural network training, we follow an iterative procedure to minimize either \eqref{eq:training} or \eqref{eq:training_inp}, stopping at the iteration where the mean squared error over all the patches $ \P_k \in \D_V $ is minimum.

Specifically, we use Adam optimization algorithm \cite{kingma2014adam}, with learning rate and patience (\red{i.e., the number of epochs with no improvement after which training will be stopped}) initialized at $0.01$ and $10$, respectively.
The former is decimated while the latter is halved in presence of plateau of the cost function.
In general, we train the network for a maximum number of $100$ epochs, although we verified the smallest loss on validation patches is often achieved within the first $30$ training epochs.

\subsection{System Deployment}
\label{subsec:deployment}

\begin{figure*}[t]
	\centering
	\includegraphics[width=\textwidth]{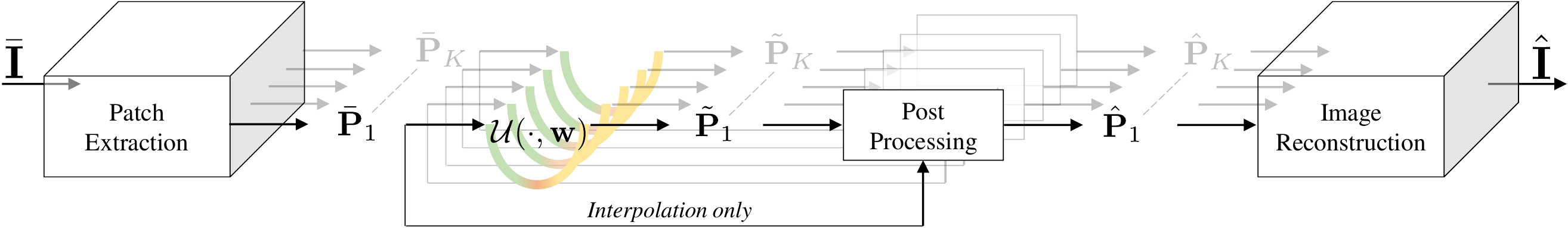}
	\caption{Proposed procedure for recovering each corrupted gather $ \Ihole $.}
	\label{fig:system_deployment}
\end{figure*}

When a new corrupted gather $\Ihole $ belonging to evaluation set $ \D_E $ is under analysis, its recovered version is estimated following the scheme depicted in Fig.~\ref{fig:system_deployment}.

First of all, a set of $K$ patches is extracted from image $\Ihole$ as described in the previous Section.
Then, each patch $\bar{\mathbf{P}}_{k}$ is processed by the \emph{U-net} architecture in order to estimate the patch $\tilde{\mathbf{P}}_{k}$.

Following the same logic of the training phase, the estimated patches are post-processed in slightly different ways according to the specific goal.
For denoising only and joint denoising/interpolation, each patch $\tilde{\mathbf{P}}_{k}$ simply undergoes a denormalization step, thus it is divided by the gain $G$ to obtain the output patch $\hat{\mathbf{P}}_{k}$.
Concerning the task of interpolation only, it is reasonable to leave the known samples untouched in the final estimated patch. 
Therefore, exploiting the binary mask defined in \eqref{eq:mask}, each patch $\hat{\mathbf{P}}_{k}$ is obtained as
\begin{equation}
\hat{\P}_{k}= \frac{\bar{\mathbf{P}}_{k} \otimes \bar{\mathbf{M}} +\tilde{\mathbf{P}}_{k} \otimes \mathbf{M}}{G},
\end{equation}
being $\bar{\mathbf{M}}$ the \red{the logical complement (i.e., the negation)} of $\mathbf{M}$.

Eventually, in order to reconstruct the image gather $\hat{\mathbf{I}}$, all the estimated patches $\hat{\P}_{k}$ are re-assembled together, sample-wise averaging the overlapping portions if some overlap between patches was used during patch extraction procedure.
\section{Results}
\label{sec:results}

In this section we present the result of our experiments, obtained on well-known public datasets.
Specifically, we evaluate our methodology over both synthetic and field data.  
First, we introduce the accuracy metrics we exploit for evaluating the results, and we show a tutorial example considering a very simple case of study.
Second, we separately validate the performances of the proposed interpolation and denoising strategies on both synthetic and \red{field} data.
Finally, we investigate the combined interpolation and denoising problem also comparing our method against a recent solution.

\subsection{Accuracy metrics}
\label{subsec:snr_metrics}

We evaluate the performances of our method in reconstructing each entire seismic image belonging to the evaluation set $ \D_E $, namely the corrupted gathers which have never been seen by the \emph{U-net}.
The accuracy metrics is the same used by \cite{jin2018seismic} and \cite{siahkoohi2018seismic}, i.e., the \red{$\mathrm{S/N}$}, defined as the ratio between the \red{power} of the original gather and the \red{power} of the reconstruction error:

\begin{equation}
\mathrm{S/N} = 10 \, \log_{10} \frac{|| \, \I \, ||^2_{\mathrm{F}}}{ || \, \I - \Ihat \, ||^2_{\mathrm{F}}} \quad  \I \in \D_E.
\label{eq:test_snr}
\end{equation}

\subsection{Tutorial example}
\label{subsubsec:tutorial_interpolation}

\begin{figure*}[t]
\centering
  \includegraphics[width=0.8\textwidth]{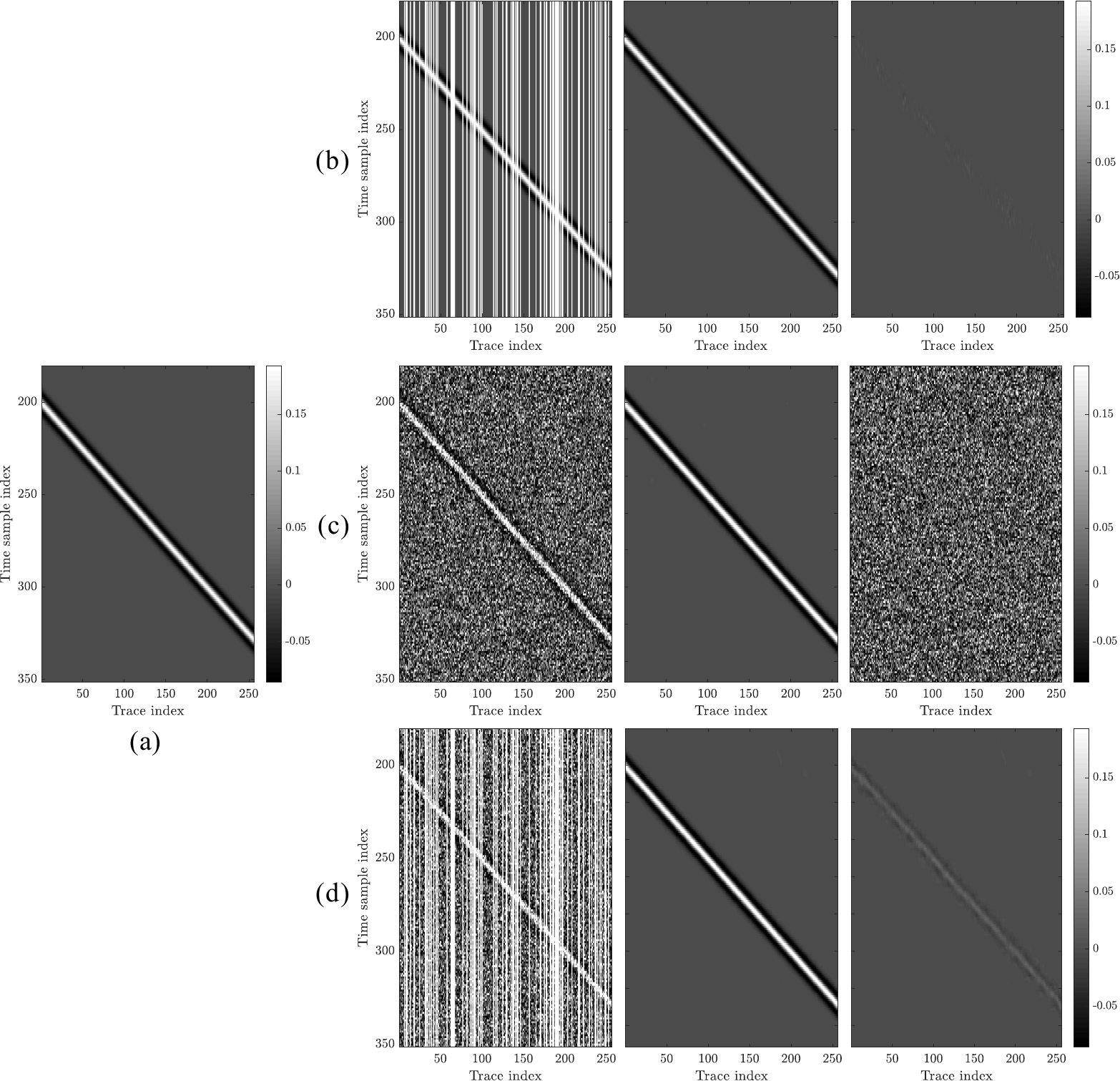}
  \caption{Tutorial example results: (a) shows the original gather; (b) depicts results for interpolation of random missing traces; (c) draws results for AWGN denoising; (d) shows results for joint interpolation and denoising. For each corruption, first column reports the corrupted gather, second column the reconstructed one. For problems (b) and (d), third column shows the reconstruction error (i.e., the difference between reconstructed and original), while for problem (c) it reports the residual error (i.e., the difference between reconstructed and corrupted).}
  \label{fig:linear_single}
\end{figure*}

In order to show the effectiveness of our strategy, we start with a very simple case of study, that is, the corrupted gather under investigation consists of a single dipping event with a Ricker wavelet.  In order to train, validate, and test the network, we build a dataset of $251$ gathers with size $512 \times 256$, randomly varying slope and depth of the event. 
In particular, we always train and validate the network using the first $250$ gathers, and we test over the last one. From each gather, we extract $153$ patches with size $128 \times 128$ and with stride $(24, 16)$ in the temporal and spatial domain, respectively.
For what concerns the rationale behind the patch selection methodology, we refer the reader to the next section, where we explain in details the motivations behind our choice.

We consider three different data corruptions: (i) randomly located missing traces; (ii) additive white Gaussian noise (AWGN); (iii) joint presence of noise and missing traces.
Concerning the first situation, we randomly delete the $30\%$ of the traces from each gather; in the second case, we add AWGN noise with standard deviation equal to $0.1$; the third example considers both corruptions. 
Results of the gather reconstruction are shown in Fig.~\ref{fig:linear_single}. The achieved S/N are $34.2\textrm{dB}$, $19.5\textrm{dB}$ and $20.1\textrm{dB}$, respectively. 
It is worth noting that, in all the three situations, we are able to reconstruct the corrupted shot gather with an acceptable level of accuracy, introducing few signal leakage. 

To compare our results with more standard solutions, we consider an algorithm based on shaping regularization for random missing traces recovery \cite{chen2015seismic} and a method based on f-x deconvolution for random noise attenuation \cite{liu2012random}.
Concerning the interpolation of missing traces, the S/N achieved by \cite{chen2015seismic} is $33.5\textrm{dB}$, thus it is comparable with our result. Regarding the denoising problem, the method based on f-x- deconvolution only achieves $\textrm{S/N} = -2.3\textrm{dB}$. 
Despite the big gap between this solution and ours, notice that the initial S/N of the gather corrupted by AWGN was $-15.8\textrm{dB}$, therefore also the f-x deconvolution strategy is able to reconstruct the gather with an acceptable accuracy. Furthermore, contrarily to our proposal which is data-driven, f-x deconvolution is totally blind and does not need training data for estimating the result.

These examples show that a properly trained \emph{U-net} is able to deal with the tasks of interpolation and denoising: the network is able to recognize the features characterizing linear events, and to effectively reject as noise all other components. 
As a matter of fact, the training set has been designed in order to incorporate prior knowledge that actual signals are linear events.
However, the main benefit of \emph{U-net} and more generally of all CNNs is that, through learning, they are in principle able to handle complex situations which cannot be easily described by standard approaches based on mathematical models of noise and seismic events.

\subsection{Problem 1: Interpolation of missing traces}
\label{subsec:interpolation}

In this section we present the network performances in interpolating gathers with missing traces. 
In particular, we start showing results achieved over a synthetic dataset, considering various interpolation situations. 
Finally, we evaluate the proposed strategy on real field data.

\subsubsection{Synthetic data \red{set}}
\label{subsubsec:synth_presentation}

The reference dataset used to systematically explore the results is extracted from the well known synthetic BP-2004 benchmark \cite{billette20052004}. In particular, we work with $1348$ shot gathers, cropped at the first $1152$ traces (taking the source as reference) and at the first $1920$ time samples/trace. The central frequency of each trace is $27\textrm{Hz}$, sampled every $ \delta_t = 6\textrm{ms}$, and the group spacing is $12.5\textrm{m}$.

\begin{figure}[t]
\centering
  \includegraphics[width=0.95\columnwidth]{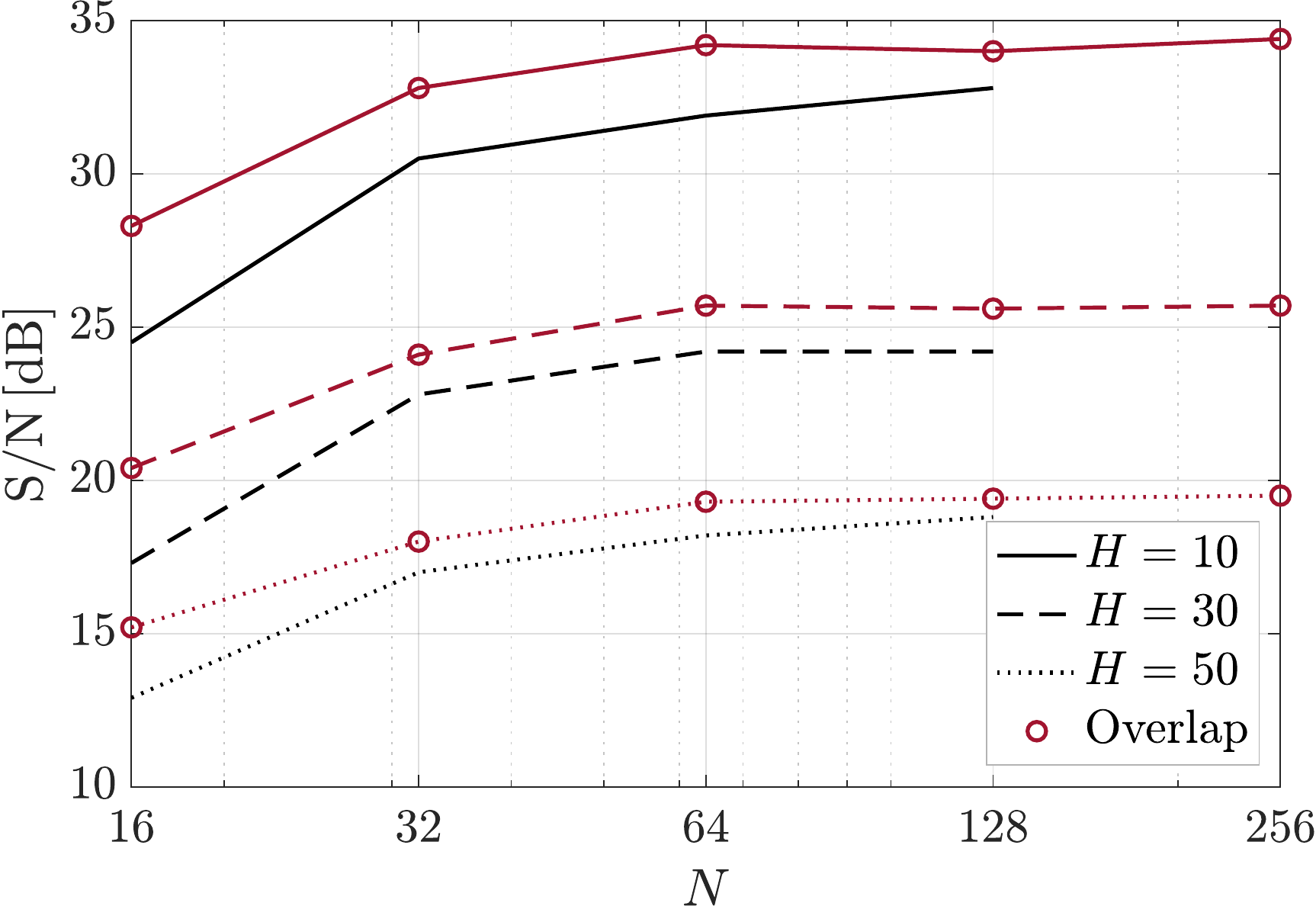}
  \caption{Average $\mathrm{S/N}$ achieved on gathers belonging to $\D_E$, as a function of patch dimension $N$ and overlap.}
  \label{fig:test_ov_patchsize}
\end{figure}

\begin{figure}[t]
\centering
  \includegraphics[width=\columnwidth]{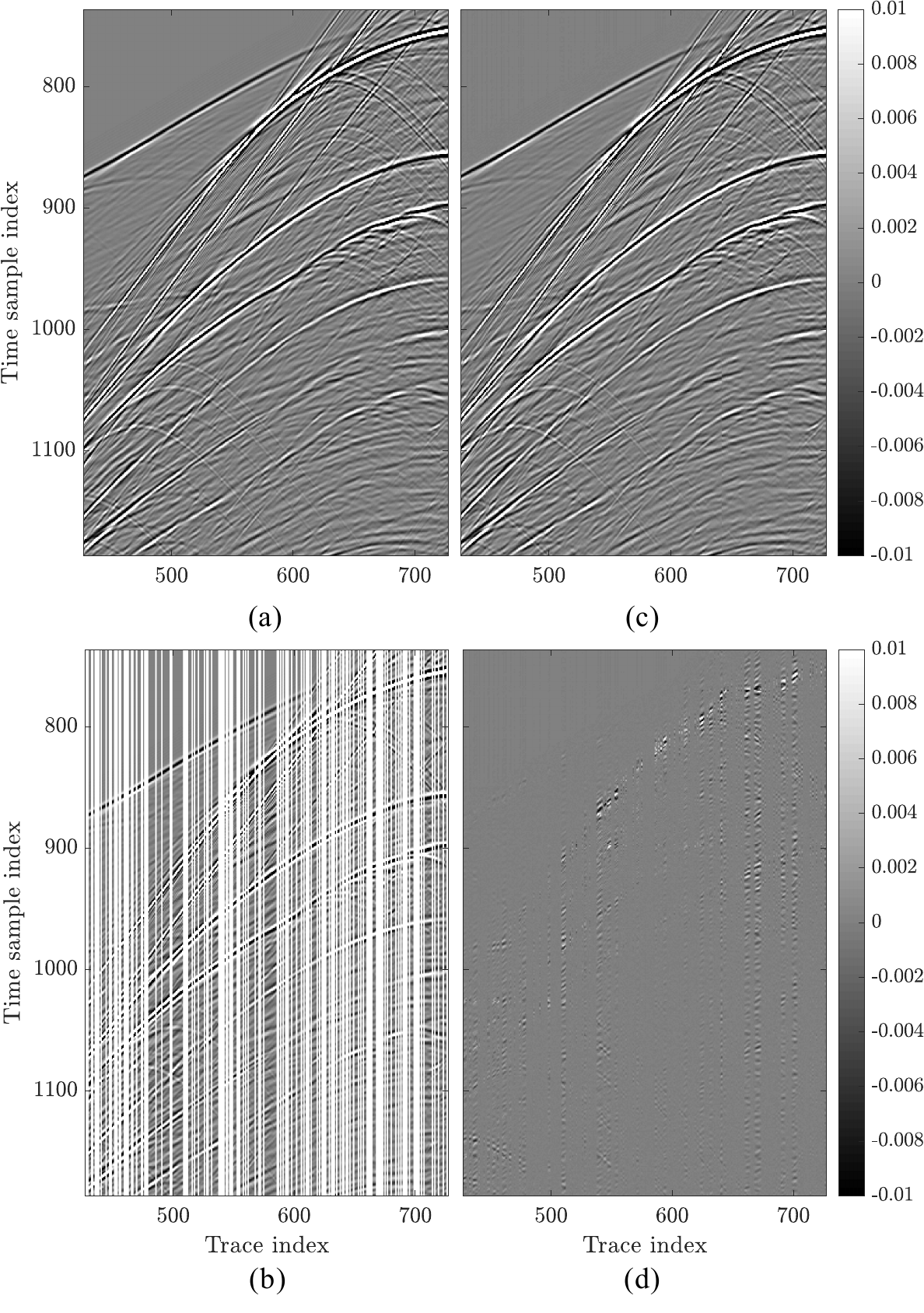}
  \caption{Example of data interpolation considering one gather of the synthetic dataset. (a) depicts the original gather $\I$, cropped in its central portion with size $450 \times 300$; (b) reports the corrupted gather $\Ihole$, with $50\%$ of randomly missing traces, (c) shows the reconstructed gather $\Ihat$; \red{(d) depicts the reconstruction error, which is the difference between reconstructed and original shot gather.}}
 \label{fig:delete050_conerr_V}
\end{figure}

In order to properly evaluate the proposed method, we randomly split the dataset into training, validation and evaluation, using $ 250 $ shot gathers for training and validation (further split on $75\%$ of images for training set $ \D_T $, and $25\%$ for validation set $ \D_V $), and the remaining for evaluation set $ \D_E $.

\paragraph{Interpolation of uniformly distributed missing traces}
\label{subsubsec:uniform_interp}
The first experiment investigates the situation of \red{randomly located} missing traces with uniform distribution. 
This choice follows the main reasoning of the works proposed in literature \cite{mandelli2018seismic,chen2015seismic,zhu2017joint}.

In order to simulate \red{a seismic acquisitions with randomly missing traces}, we extract $3$ different datasets from the reference one, deleting a percentage $ H $ of the available data traces. To be precise, for each shot gather $ \I$, we randomly delete the $ H \% $ of its traces, $ H \in \{10, 30, 50 \} $, obtaining a holed gather $ \Ihole $.

As shown in Section \ref{subsec:unet_implementation}, we work in patch-wise fashion for reconstructing the corrupted gathers. 
Specifically, each gather entering the network is initially split into a plurality of squared patches, with dimensions $N \times N$. 
In light of this, we perform an initial experiment to analyze the behaviour of network output as a function of the specific input data. 
The goal of this primary investigation is to select a good patch extraction method, that is, the strategy leading to the highest reconstruction accuracy on the evaluation set. 
We consider different values for $N$, namely $N \in \{ 16, 32, 64, 128, 256 \}$, and we evaluate the cases of non-overlapping patches and of patches extracted with an overlap of $N / 2$ in both directions. 

To evaluate the $\emph{U-net}$ performances according to the chosen patch extraction method, we use $\mathrm{S/N}$ defined in \eqref{eq:test_snr}.
Fig.~\ref{fig:test_ov_patchsize} shows the average \red{$\mathrm{S/N}$} achieved over gathers belonging to evaluation set, with and without the overlap between the extracted patches. Note that the case $N = 256$ does not include results without overlap because the gather dimensions are not integer multiples of this value.
It is noticeable that small values of $N$ are not good solutions for reconstructing the corrupted images, probably because the \emph{U-net} needs to analyze more samples together in order to find a \red{significant hidden} representation of the input patch. 
As expected, introducing some overlap during patch extraction always returns better performances than just selecting adjacent patches. 
This is due to two main factors: 
first, selecting overlapped patches increases the amount of data seen by the network and reasonably improves its performances; second, in the image reconstruction phase, the overlapping portions of the patches are sample-wise averaged, decreasing the possibility to generate undesired edge/border effects.

Even if selecting an overlap of $N/2$ gives slightly better results, one consideration must be done.
At training stage, we found out that a good strategy is to group in a single batch all the patches extracted from the same shot gather, ending up with a batch size (i.e., the amount of \emph{patches} in a batch) strictly dependent on $N$. 
Notice that the number of \emph{samples} per batch does not change with $N$, if patches are not overlapped. Conversely, in case of overlapped patches, the number of samples per batch increases, as some samples belong to multiple patches.
Therefore, the higher the overlap, the larger the amount of GPU memory required in training phase.
If the absence of overlap requires a GPU memory usage more or less equal to $4\textrm{GB}$ for every $N$, in case of overlap the required space increases in a quadratic fashion.

Therefore, considering that the achieved \red{$\mathrm{S/N}$} performances of the two methodologies (overlapped and non-overlapped patches) are not so far, we choose the patch extraction strategy which selects only adjacent and non-overlapping patches. 
For this reason, hereinafter we only investigate the network behavior considering non-overlapp\red{ing} patches, as overlapping \red{patches} would make the solution impractical in the majority of cases.

Regarding the patch dimension $N$, as the \red{$\mathrm{S/N}$} curve monotonically increases with the patch dimension but without dropping performances in terms of memory usage, we select $ N = 128$ for all the \red{experiments}.
We end up with batches of $135 $ non overlapping patches with dimensions $128 \times 128$ extracted from each shot gather. 
The process involves more than $ 25\,000 $ training patches, more than $ 8\,500 $ validation patches, and more than $ 145\,000 $ testing patches for each dataset.

Regarding the results, we are able to achieve \red{$\mathrm{S/N}$} of $32.8\textrm{dB}$, $24.2\textrm{dB}$ and $18.8\textrm{dB}$ for $H = 10, 30$ and $50$, respectively. 
The processed gathers do not visually show any artifacts due to the interpolation method, even with $ H = 50$, as depicted by the example in Fig.~\ref{fig:delete050_conerr_V}. 
\red{For what concerns the required computational time, we need more or less $75$ minutes for training and validate the network over $100$ epochs. Despite this could seem a quite huge amount of time, once training has been done, we are able to recover a corrupted shot gather belonging to the test set in less than $0.3$ seconds. Specifically, we run our tests on a Workstation equipped with Intel Xeon E5-2687W v4 (48 Cores @ 3 GHz), RAM 252 GB and 1 TITAN V (5120 CUDA Cores @ 1455MHz), 12 GB.}

\paragraph{Interpolation of bursts of missing traces}
\label{subsubsec:markov}

Uniform distribution of missing traces, described by the percentage $H$, allows the evaluation of average reconstruction performances of the interpolation. However, in order to have a more detailed description, here we study the performances of the proposed \emph{U-net}-based interpolation on a more sophisticated corruption model. Basing on the consideration that missing traces (due for instance to spatial obstacles) are likely to appear in groups, we propose a \red{bursty} missing traces model inspired by the packet loss models of telecommunication networks \cite{milner2004analysis}. \red{The term \emph{burst} is taken from the field of telecommunication networks, and refers to groups of consecutive events (in this case, groups of consecutive missing traces). The more bursty a distribution of missing traces is, the more the missing traces are likely to cluster.}

In particular, the model is a two states Markov model described by two parameters, $\alpha$ and $\beta$: $\alpha$ refers to the probability of a missing trace, while $\beta$ is the average length of the burst, i.e., the average number of missing traces which are adjacent one another. 

\begin{figure}
	\centering	
	\includegraphics[width=0.65\columnwidth]{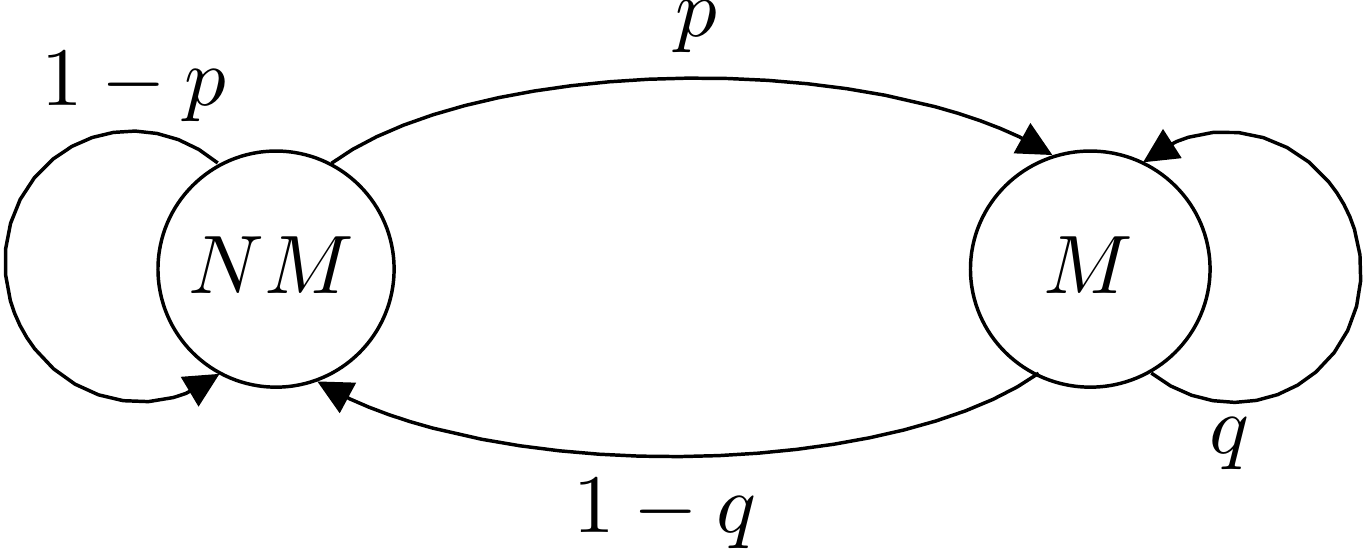}
	\caption{Markov chain of the burst corruption model. }
	\label{fig:markov}
\end{figure}

The Markov chain of the model is depicted in Fig.~\ref{fig:markov}, where $NM$ represents the non-missing trace state whereas $M$ is the state for missing trace. The probability to find a corrupted trace, given that the previous one (in the spatial dimension) was missing, is $q$, while the probability to pass from a non missing trace to a missing one is $p$. These probability values can be derived from $\alpha$ and $\beta$, formally, 
\begin{equation}
q =  1 - \frac{1}{\beta}, \quad
p=  \frac{\alpha}{\beta(1-\alpha)}.
\end{equation}

Exploiting this model, we can simulate more realistic scenarios, where \red{clusters} of adjacent missing traces can occur, due for instance to environmental constraints or sudden interruptions during acquisitions.   
In order to test our method on this missing trace distribution, we select various percentage of missing traces $\alpha \in \{10, 30, 50\} \%$ with average burst length $\beta \in \{ 1, 2, 3 \}$, corresponding to $12.5\textrm{m} , 25\textrm{m}$ and $37.5\textrm{m}$ respectively. Notice that, the larger the average gap, the greater the gap size dispersion. Indeed, for $\beta=1$ the standard deviation is equal to $0$ traces (isolated missing traces only); on the contrary, for $\beta=2$ and $\beta=3$ the standard deviations are $\sigma=1.14$ and $\sigma=2.44$ traces, respectively. For instance, in the datasets under examination, $\beta=3$ provides a maximum gap up to $30$ traces (corresponding to $375\textrm{m}$), which simulates a quite large physical obstacle. 

\begin{table}[t]
	\caption{Average $\mathrm{S/N}$ [dB] achieved on gathers belonging to $\D_E$, for different values of $\alpha$ and $\beta$.}
	\label{tbl:markov}
	\centering
	\def\arraystretch{1.6}
	\resizebox{0.48\columnwidth}{!}{
\begin{tabular}{c|*{3}c}
			\diagbox[width=1.5cm, height=0.5cm]{\raisebox{.5pt}{ $\alpha$} }{\raisebox{5pt}{ $\beta$} }
			& $ 1 $    & $2 $     &  $ 3$    	 	\\  \Xhline{2\arrayrulewidth}
			$10$  & $38.4$ & $25.3 $ & $21.9$ \\ \hline
			$ 30 $ & $ 32.8$ & $21.7$ & $18.1 $ \\ \hline
			$ 50$ & $29.9$ & $18.7$ & $15.7$ \\
\end{tabular}}
	\end{table}

Table \ref{tbl:markov} depicts the average results achieved by the \emph{U-net} on the evaluation set. 
Notice that, the larger the burst length, the lower the resulting \red{$\mathrm{S/N}$}. 
This enlightens the need of further investigations for interpolating bursts of many traces: as a matter of fact, as the \red{size of the} group of adjacent missing traces increases, the ability of the network in reconstructing the unknown samples diminishes.
Nonetheless, notice that even in the worst case, i.e., $(\alpha, \beta) = (50, 3)$, the \emph{U-net} is able to maintain acceptable reconstruction performances. 

\paragraph{Interpolation by transfer learning}
\label{subsubsec:ts_diff}

\begin{figure}[t]
	\centering	
	\includegraphics[width=\columnwidth]{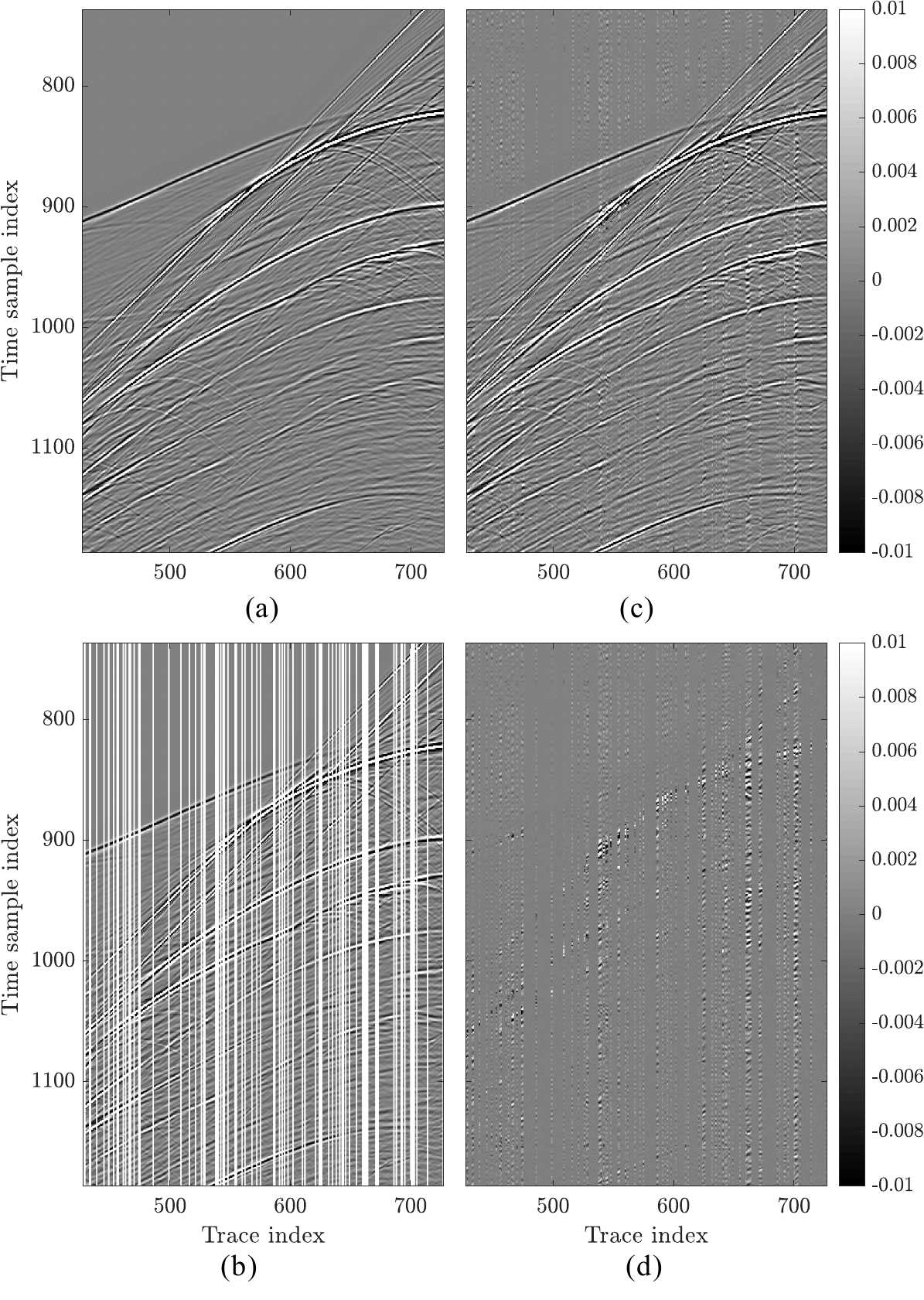}
	\caption{Example of transfer learning data interpolation. (a) depicts the original gather $\I$, sampled every $8\textrm{ms}$ and cropped in its central portion with size $450 \times 300$; (b) shows the corrupted gather, with $30\%$ of randomly distributed missing traces; (c) reports the reconstructed gather $\Ihat$, exploiting the \emph{U-net} trained on data sampled every $6\textrm{ms}$; (d) shows the reconstruction error, i.e., (c) - (a). }
	\label{fig:6ms_on_8ms_conerr_V}
\end{figure}

In order to test the robustness of the proposed method in interpolating missing data, we generate two further synthetic datasets, exploiting the very same acquisition geometry and model of the dataset previously presented, but with different sampling rates $4\textrm{ms}$ and $8\textrm{ms}$. 
The goal of this experiment is to check if the \emph{U-net} architecture, when trained on data sampled every $6\textrm{ms}$, is able to reconstruct differently-sampled data. This is an example of the well-known transfer learning strategy \cite{goodfellow2016}. Namely, it corresponds to analyzing the performances of one network which has already been trained over a dataset having different features from the testing one. 

To this purpose, we propose to select as test case the uniform missing traces framework, randomly deleting the $30 \%$ of traces from these new datasets. 
Then, we evaluate the reconstruction results on gathers belonging to the evaluation set $\D_E$ of these datasets, with the difference that we exploit the network trained on the dataset sampled every $ \delta_t = 6\textrm{ms}$. 

Average results of the interpolation are shown in Table~\ref{tbl:ts_diff}. 
Notice that we report also the interpolation results we can achieve if following the standard training \red{procedure}, that is, training the network using data with the same sampling time of the evaluation set.
Even if the difference between the results is noticeable, Fig.~\ref{fig:6ms_on_8ms_conerr_V} shows an example of $\delta_t = 8\textrm{ms}$ data reconstruction exploiting the \emph{U-net} trained on $\delta_t = 6\textrm{ms}$. 
\red{It is noticeable that the error is concentrated in the upper part of the gather and in the areas with many adjacent missing traces, while in the other regions the reconstruction is quite good and acceptable.}
\begin{table}[t]
	\caption{Average $\mathrm{S/N}$ [dB] achieved on $\D_E$, for sampling time $\delta_t = 4\textrm{ms}$ and $\delta_t = 8\textrm{ms}$.}
	\label{tbl:ts_diff}
	\centering
	\def\arraystretch{1.6}
	\resizebox{0.52\columnwidth}{!}{
 \begin{tabular}{c?cc}
	       $\delta_t$ & 
       $ 4\textrm{ms}$ & 
       $ 8\textrm{ms}$ \\
       \Xhline{2\arrayrulewidth}
       Train on $6\textrm{ms}$ &
       $ 10.1 $ &
       $ 10 $ \\ \hline
       Train on $\delta_t$ &
       $ 25.3 $ &
       $23.8 $ \\
     \end{tabular}}
	\end{table}

\paragraph{Interpolation of regularly missing traces}
\label{subsubsec:regular_interp}
The last case we investigate for the synthetic dataset is that of recovering regularly missing traces, which can be considered as upsampling the acquisition geometry and has several practical implications, e.g., interpolation in the cross-line direction. 
This case has its own peculiarities, and it is often the most challenging interpolation problem, especially when the dips are aliased. As a matter of fact, methods based on sparse transformations and low-rank constraints have limited application in this case because of the strong spatial aliased energies. 
We want to test the conjecture that network training extracts some high-level characteristics of seismic data which are more robust to alias than linear events or low rank assumptions, therefore implicitly including an anti-aliasing strategy.

\begin{figure*}[t]
	\centering	
	\includegraphics[width=0.8\textwidth]{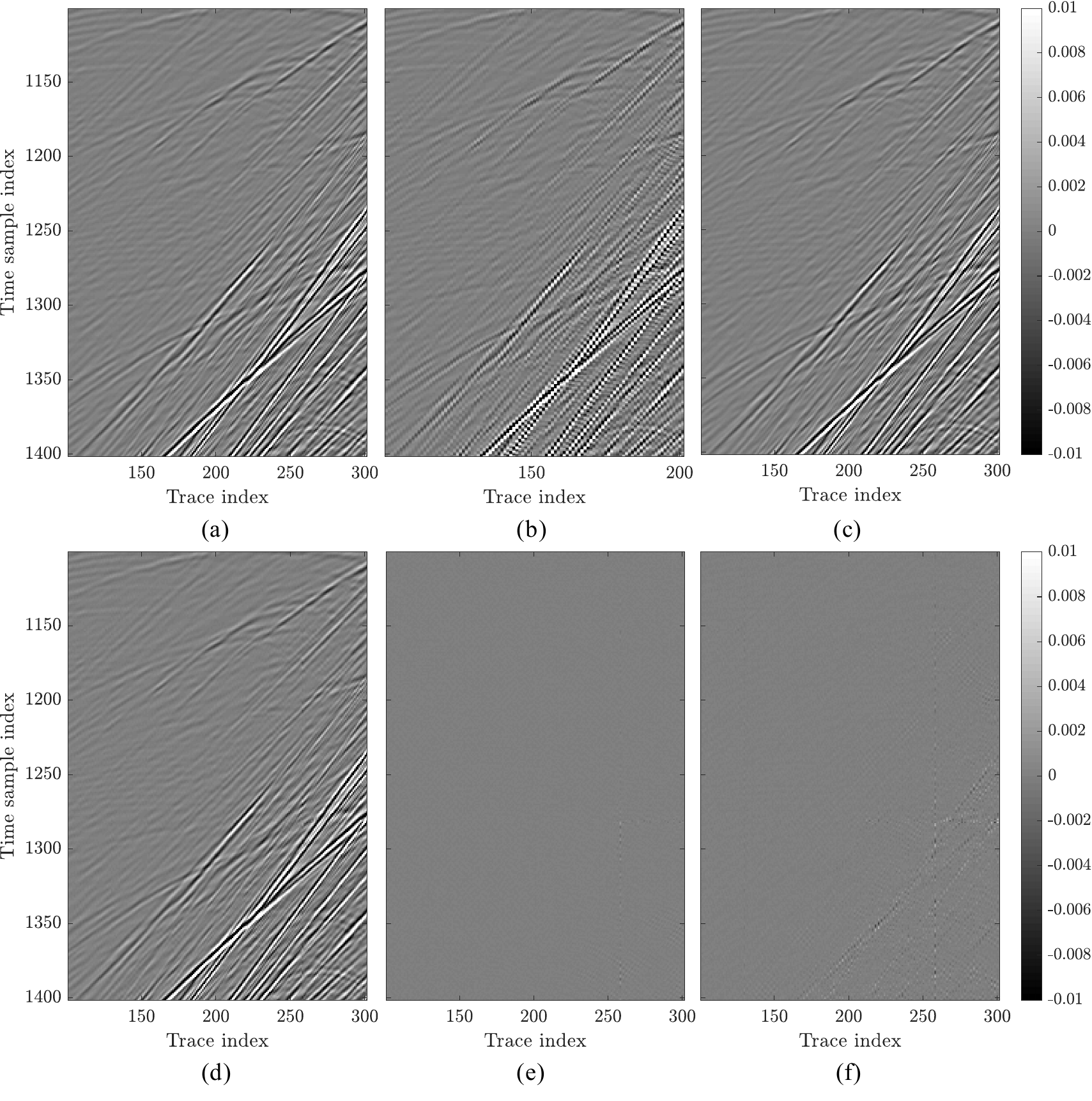}
	\caption{Interpolation of regularly missing traces. (a) depicts the original gather, cropped around samples (1100$\,$ \textendash $\,$1400, 100$\,$ \textendash $\,$300); (b) shows the corrupted gather which contains half the original traces; (c) depicts the reconstructed gather by training the \emph{U-net} on data with regularly missing traces; (d) reports the reconstructed gather by transfer learning; (e) shows the reconstruction error of the standard strategy, i.e., (c) - (a); (f) is the reconstruction error of the transfer learning approach, i.e., (d) - (a).}
	\label{fig:deleteregular050_nonan}
\end{figure*}

\begin{figure}[!]
	\centering	
	\includegraphics[width=\columnwidth]{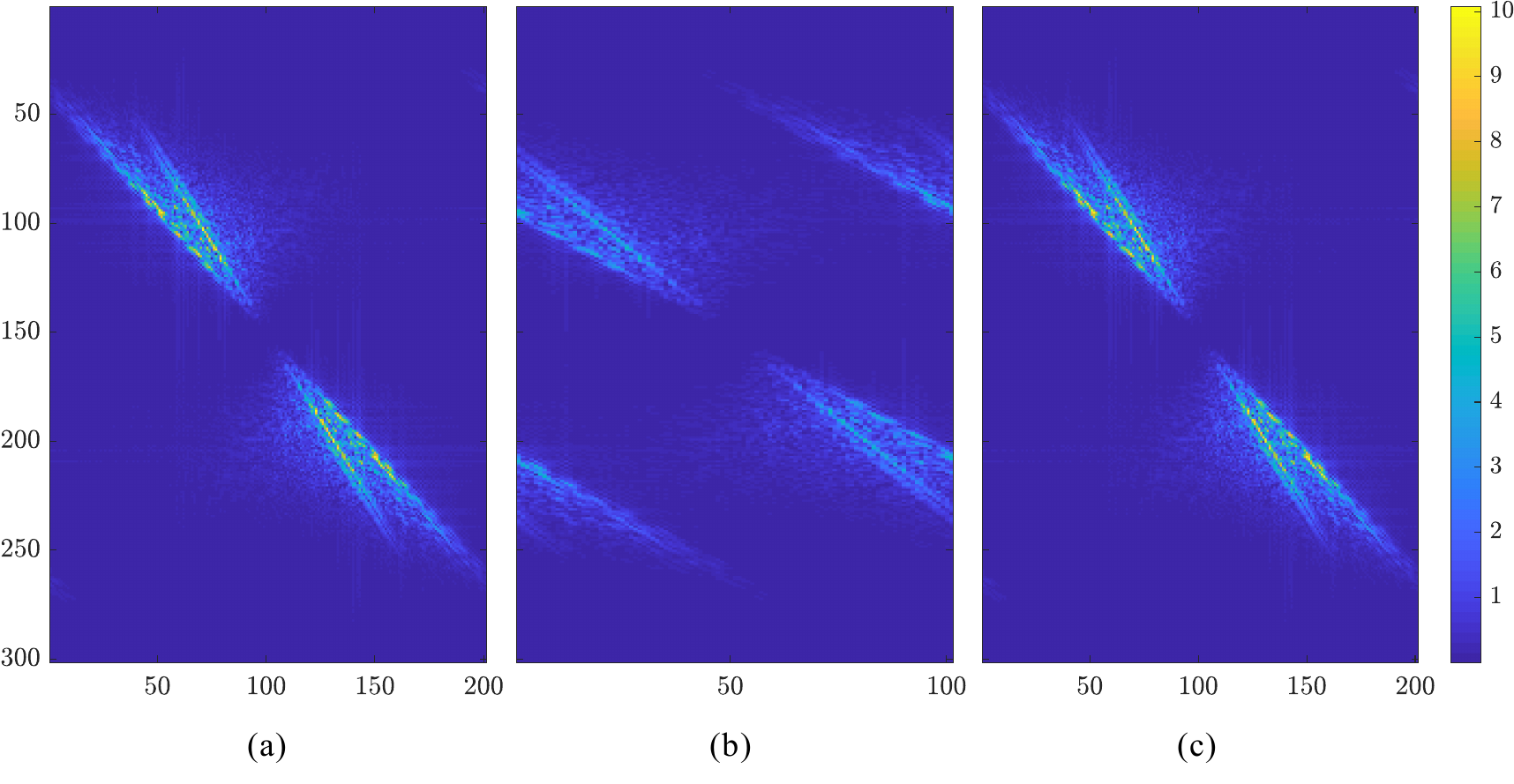}
	\caption{Absolute value of Fourier spectrum of shot gathers for the problem of interpolation of regularly missing traces. (a) depicts the spectrum of the original gather cropped around samples (1100$\,$ \textendash $\,$1400, 100$\,$ \textendash $\,$300); (b) shows the spectrum of the corrupted gather which contains half the original traces; (c) depicts the spectrum of the reconstructed gather by training the \emph{U-net} on data with regularly missing traces.}
	\label{fig:fourier_deleteregular}
\end{figure}

We consider two scenarios. 
The first one consists in training the \emph{U-net} with new data, generated by regularly deleting the $50\%$ of the shot gathers traces, and then following the same procedure previously shown for the shot gather reconstruction. 
The second solution emulates the transfer learning technique presented above: we use the network trained on data with $50\%$ randomly missing traces to reconstruct the shot gathers with $50\%$ regularly missing traces.
The achieved S/N are $30 \textrm{dB}$ and $22.8 \textrm{dB}$ for the first and second solution, respectively. 
Results achieved over a shot gather region characterized by steep dips are shown in Fig. \ref{fig:deleteregular050_nonan}.
Notice that, by training and testing the \emph{U-net} with the same data, the reconstruction error is very low, while with the transfer learning technique the error contains again some signal content. Nonetheless, this is an expected results and both errors seem acceptable at visual inspection. 
Indeed, we select as a reference a recent industrial software based on f-x deconvolution which does not require training data (hence can be considered more or less analogous to the transfer learning technique), and it achieves $\textrm{S/N} = 22.7\textrm{dB}$ on the same data. 
For the sake of clarity, Fig. \ref{fig:fourier_deleteregular} includes the absolute value of the Fourier spectrum computed for the original shot gather, for the corrupted one, and for the estimated one by means of the first proposed strategy. It is worth noting that the alias introduced in the corrupted shot gather can be deleted by our reconstruction method. 

\color{black}
\subsubsection{Field data}
\label{subsubsec:viking}

\begin{figure}
	\centering	
	\includegraphics[width=\columnwidth]{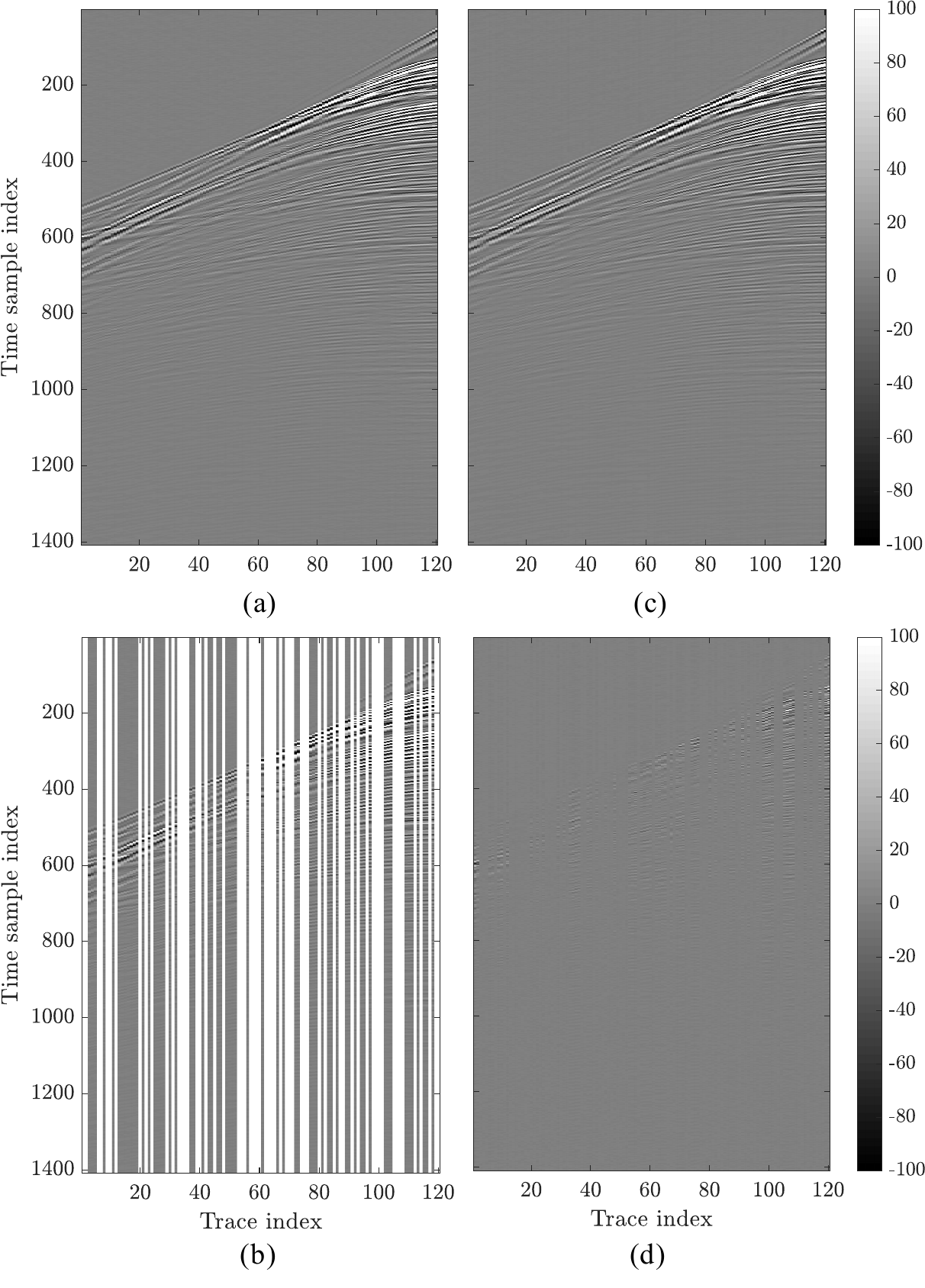}
	\caption{Example of data interpolation, considering one gather of the field dataset \cite{keys1998data}. (a) depicts the original gather $\I$; (b) reports the corrupted gather $\Ihole$, with $50\%$ of randomly missing traces; (c) shows the reconstructed gather $\Ihat$; (d) reports the reconstruction error, i.e., (c) - (a).}
	\label{fig:viking50_conerr_V}
\end{figure}

In this section, we apply the \emph{U-net} for reconstructing corrupted real seismic data \textcolor{black}{and compare the result achieved with those obtained in \cite{mandelli2018seismic}}.
To this purpose, we exploit as field data the well known Mobil Avo Viking Graeben Line 12 dataset \cite{keys1998data}. 
Specifically, this dataset consists of $ 1001 $ marine shot gathers. Each gather is composed of $128$ traces of $1408$ time samples, with temporal sampling of $ 4 \textrm{ms} $ and receiver sampling of $ 25 \textrm{m} $.

In order to compare our results with those obtained by \cite{mandelli2018seismic}, we simulate \red{ a seismic acquisition with a uniform distribution of randomly missing traces}.
Therefore, for each acquired gather $ \I$, we randomly delete the $ H \% $ of its traces, $ H \in \{10, 30, 50 \} $, obtaining a scattered sampled gather $ \Ihole $. 

Following the same rationale of the synthetic example, we split each dataset into $250$ gathers for training and validation and leave the remaining to evaluation set. Then, in order to achieve a similar number of patches per gather (i.e., $135$ in the synthetic case), we extract $129$ patches with size $128 \times 128 $, overlapped \red{only on the} temporal dimension, specifically with patch-stride of $10$ samples. 
Notice that, in this case, the presence of patch overlap does not cause issues in memory usage, as the number of samples entering the network is similar to the chosen configuration for the synthetic example. 

Results obtained on the evaluation dataset $ \D_E $ are reported in Table \ref{tbl:field_res}, while Fig.~\ref{fig:viking50_conerr_V} shows an example of gather reconstruction where $50 \%$ of traces is missing.
It is noticeable the improvement in performances of the proposed architecture, 
This achievement is due to the specific changes performed on the \emph{U-net} architecture as described in Section \ref{subsec:unet_implementation}. As a matter of fact, even \red{with} a reduced amount of gathers for training and validation (i.e., $25 \%$ of the whole dataset instead of $75 \%$), the resulting $\mathrm{S/N}$ always exceeds the past performances.

\begin{table}[t]
	\caption{Average $\mathrm{S/N}$ [dB] achieved on gathers from dataset in \cite{keys1998data}. }
	\label{tbl:field_res}
	\centering
	\def\arraystretch{1.6}
	\resizebox{0.58\columnwidth}{!}{
\begin{tabular}{c?ccc}
	       $H$ & 
       $ 10 $ & 
       $ 30 $ & 
       $ 50 $ \\
       \Xhline{2\arrayrulewidth}
       Proposed \emph{U-net} &
       $ 25.7 $ &
       $20.5$ &
	   $ 16.7 $ \\ \hline
  \cite{mandelli2018seismic}&
       $ 22.5 $ &
       $15.1$ &
	   $ 10.2 $ \\
     \end{tabular} }
	\end{table}

\begin{figure}
	\centering	
	\includegraphics[width=\columnwidth]{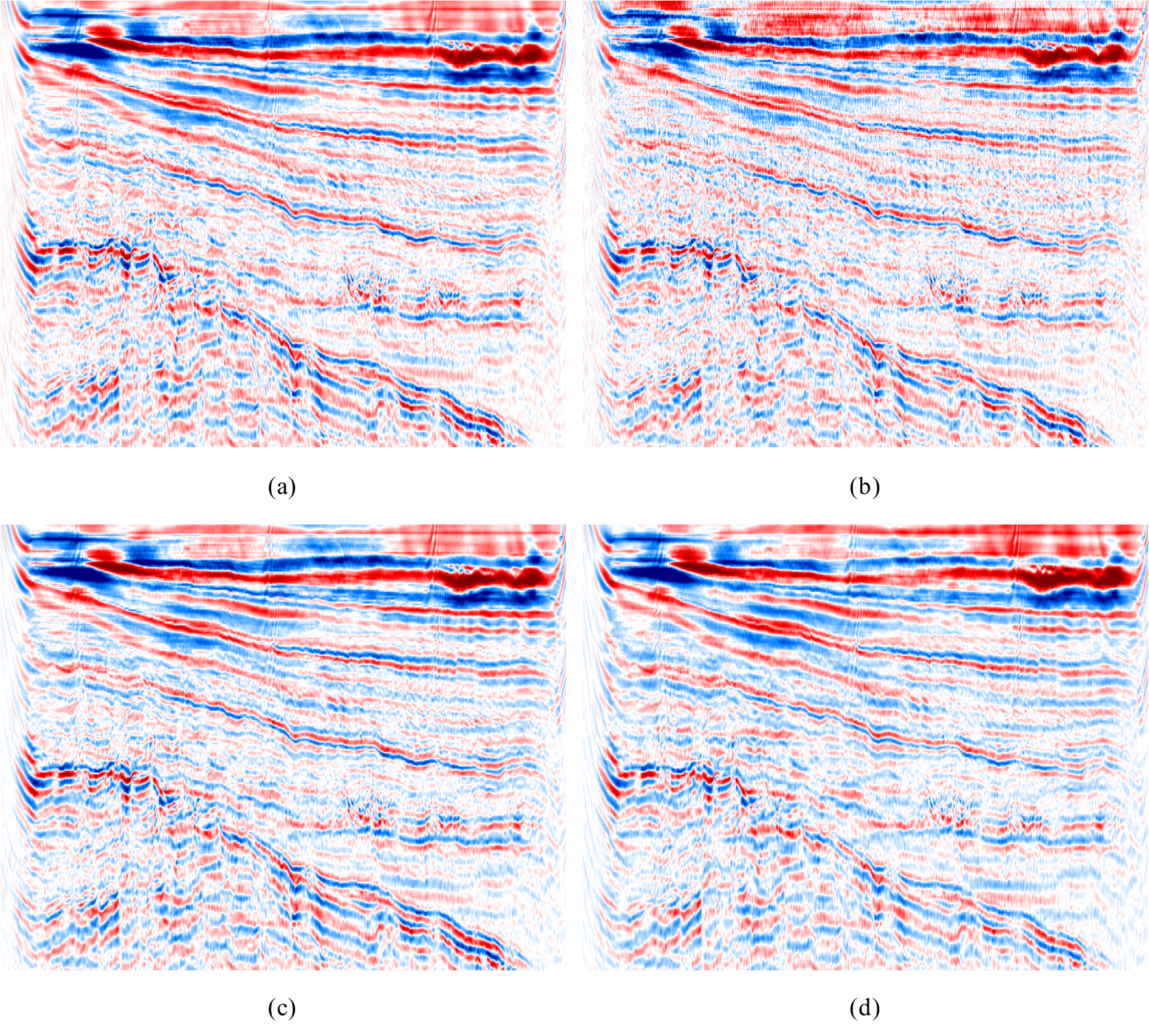}
	\caption{Kirchhoff migrated section of the Viking Graben dataset: (a); migration of original data; (b) migration of corrupted data; (c) migration of data reconstructed by \emph{U-net}; (d) migration of data reconstructed by the method proposed in \cite{chen2015seismic}.}
	\label{fig:migrated_viking_conshaping}
\end{figure}

\red{In order to build a preliminary evaluation of the results which would be obtained by seismic imaging algorithms on data reconstructed by \emph{U-net}, we migrate the sections of the original Viking Graben dataset, of the corrupted dataset with $50 \%$ randomly missing traces and of the dataset reconstructed by the proposed methodology. 
As a comparison, we migrate also the dataset reconstructed using the nonlinear shaping regularization method proposed by \cite{chen2015seismic}. 
The images depicted in Fig.~\ref{fig:migrated_viking_conshaping} show that, in case of data reconstructed with \emph{U-net}, noise is attenuated and loss of continuity is almost perfectly recovered in the migrated section. This occurs also for the reconstruction of \cite{chen2015seismic}, 
however, contrarily to that of \emph{U-net}, the migrated section shows some spurious events which are not present in the original image.}

\subsection{Problem 2: Denoising of corrupted gathers}
\label{subsec:denoising}

In this section we report the results of the \red{numerical experiments} related to denoising of corrupted gathers. We consider two kinds of additive noise, the standard \red{a}dditive \red{w}hite Gaussian \red{n}oise (AWGN), and a spike-like noise.

Concerning the patch extraction methodology to be applied, we select exactly the same strategy of that presented in Section \ref{subsubsec:synth_presentation}. 
As a matter of fact, the presence of randomly missing traces as well as the additive random noise can be seen as two generic kinds of gather corruption, which can be tackled by the \emph{U-net} in a similar way.

\subsubsection{AWGN noise model}
\label{subsubsec:awgn}

 In order to test our method over a plurality of signal to noise ratios (\red{$\mathrm{S/N}$}), we add white gaussian noise for achieving $\red{\mathrm{S/N}} = S \in \{-3, 0, 3 \}\textrm{dB}$, defined as the ratio between the \red{signal} and noise \red{power}.

The average results obtained on shot gathers belonging to the evaluation set $ \D_E $ are the following: $\red{\mathrm{S/N}} = 12.8\textrm{dB}$, $14.4\textrm{dB}$ and $16.3\textrm{dB}$, correspondent to increasing values of $S$.
Indeed, these results can be considered an upper bound for the achievable performances of the proposed strategy in realistic scenarios. As a matter of fact, they are obtained on the assumption to have clean data available for the training phase, which is never the case for field acquired data.

\subsubsection{Spike-like noise model}
\label{subsubsec:saltpepper}
 Pre-stack seismic data can be affected by different types of random noise coming from various sources, such as wind motion, poorly planted geophones or electrical noise, most of these being far more complex than simple AWGN. 
For instance, some of these seismic noises exhibit spike-like characteristics \cite{liu20081d} and are lately gaining growing interest, as they strongly affect the processing of simultaneous source data acquired from recent seismic surveys \cite{zhou2017spike}. 

Therefore, we propose to use our network for denoising data corrupted by additive spike-like noise. In order to simulate this noise, we add \red{spiky} noise with variable density $d \%$, namely the percentage of corrupted samples in one gather.
In particular, the binary values of this noise are set to the minimum and maximum values of the original uncorrupted data. 
Then, we convolve each noise trace with a Ricker wavelet having the same central frequency of the data (i.e., $27\textrm{Hz}$) and unit energy. This way, we generate two corrupted datasets, corresponding to $d \in \{ 1, 3 \} $.

\begin{figure}
	\centering	
	\includegraphics[width=\columnwidth]{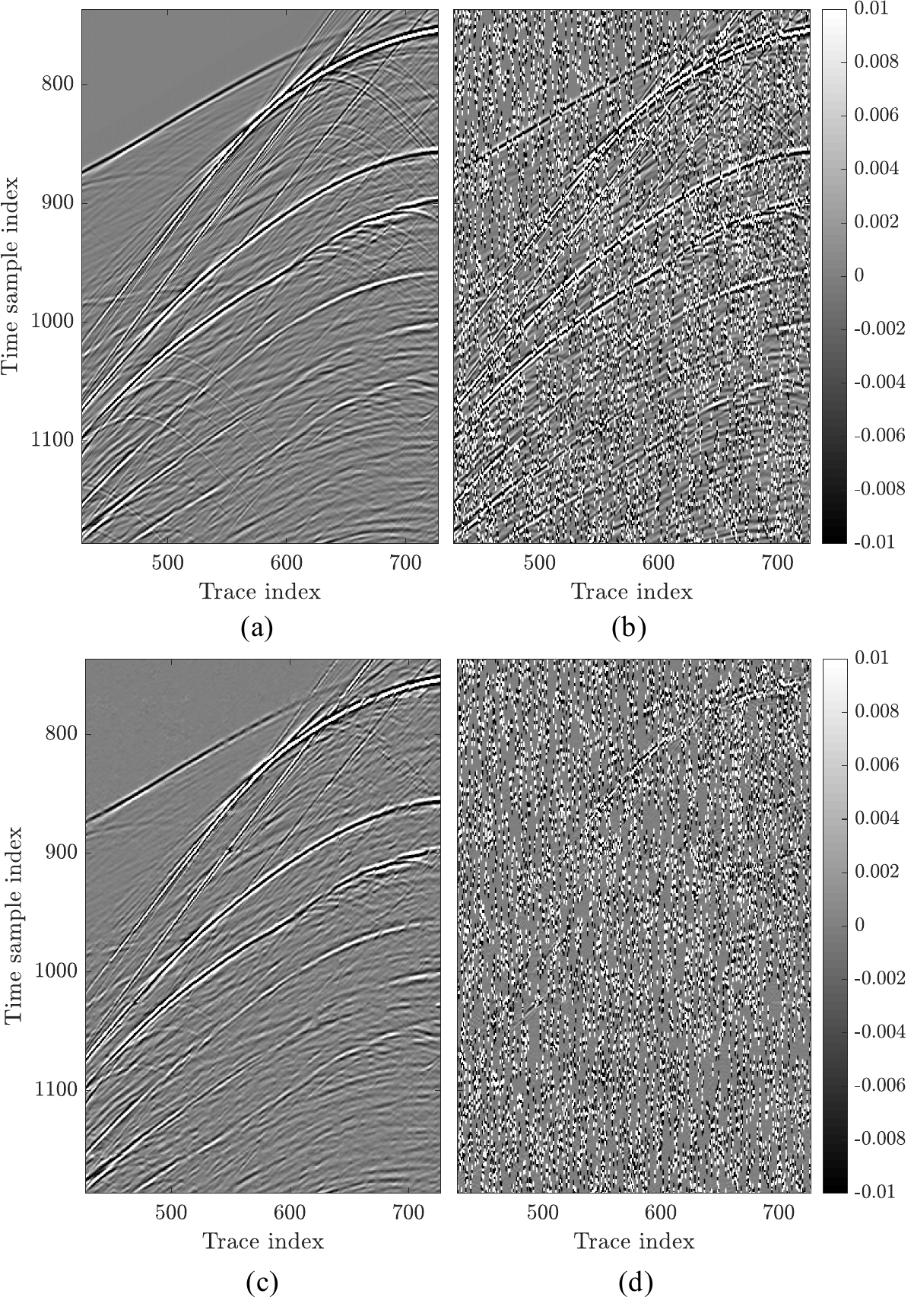}
	\caption{Example of spike-like noise attenuation. (a) reports the original gather; (b) reports the corrupted gather, with noise density $d = 3$; (c) shows the reconstructed gather; (c) shows the residual error, i.e., (b) - (c).}
	\label{fig:salt_pepper003_conres}
\end{figure}

Fig.~\ref{fig:salt_pepper003_conres} shows an example of spike-like corruption denoising for $d = 3$. It is noticeable that, even if the corrupted image visually undergoes a strong degradation, the reconstructed one presents almost all the features of the original data. \red{Moreover, Fig.~\ref{fig:salt_pepper003_conres}(d) reports the residual error between corrupted and reconstructed gather. It is worth noting that a large portion of the useful signal remains unaltered, except for very reduced areas.} 
This trend is confirmed by Table \ref{tbl:salt_pepper}, which reports the average results achieved on set $\D_E$. 

\begin{table}
	\caption{Initial $\mathrm{S/N}$ [dB] on corrupted gathers and final $\mathrm{S/N}$ [dB] achieved by spike-like denoising.}
	\label{tbl:salt_pepper}
	\centering
		\def\arraystretch{1.6}
	\resizebox{0.45\columnwidth}{!}{
   \begin{tabular}{c?cc}
	       $d$ & 
       $ 1$ & 
       $ 3$ \\
       \Xhline{2\arrayrulewidth}
       Initial &
       $ -34.8 $ &
       $-39.6 $ \\ \hline
       Final &
       $ 16.8 $ &
       $12.3 $ \\
     \end{tabular} }
\end{table}

\subsubsection{Towards standard denoiser emulation}
\label{subsubsec:wavelet_wiener}

 In order to highlight the \emph{U-net}-based method versatility, we propose to exploit our denoising strategy as an emulator of some well known standard denoising algorithms.
In particular, we select two noise attenuation strategies, namely \red{a} Wavelet \red{based} denoising \cite{jones2001scipy} and a \red{f-x deconvolution method included in a recent production software}.
To test the denoising performances, we use the datasets corrupted by AWGN with $\red{\mathrm{S/N}} = S \in \{ -3, 0, 3 \}\textrm{dB}$.

Initially, we process the whole datasets through the aforementioned standard denoising algorithms. 
Through this operation, we are generating denoised data which can be more or less considered in the same way as clean uncorrupted gathers. 

In a second phase, we train our network in a slightly different way than the approach shown Section \ref{subsec:training}. Indeed, we train the \emph{U-net} substituting to ground truth gathers those obtained through denoising by Wavelet or by \red{the industrial f-x deconvolution}.
Thus, the training step includes pairs of noisy gathers and gathers denoised by standard algorithms.
Eventually, we evaluate denoising results on shot gathers belonging to $\D_E$, comparing reconstructed gathers with the original ones, as described in \eqref{eq:test_snr}. Specifically, we compute the average results for \emph{U-net} and for standard denoising algorithms as well.

From results depicted in Table \ref{tbl:denoising} it is quite evident that performances of \emph{U-net} are comparable with those achieved through the denoising algorithm used for training, showing that \emph{U-net} is able to mimic their performances.

Moreover, the proposed method has a further advantage, which is the low computational effort in denoising a generic gather. As a matter of fact, if a \red{limited} amount of time is needed for training the network model parameters, the evaluation phase is very efficient: \red{the optimized production solution} and the Wavelet denoising
take respectively the same time and $25 \%$ more than the time required by our strategy (i.e., approximately $0.3$ seconds) for estimating each denoised gather.

\begin{table}[t]
	\caption{Average $\mathrm{S/N}$ [dB] on gathers belonging to $\D_E$, training the \emph{U-net} on gathers denoised by Wavelet (left table) and by the industrial software based on f-x deconvolution (right table).}
	\label{tbl:denoising}
\resizebox{\columnwidth}{!}{
\begin{minipage}{.22\textwidth}
\centering
\def\arraystretch{1.6}
 \begin{tabular}{c?ccc}
   $S$ & 
$ -3 $ & 
$ 0 $ & 
$ 3 $ \\
\Xhline{2\arrayrulewidth}
\emph{U-net} &
$ 4.8 $ &
$5.7$ &
$ 6.9$ \\ \hline
Wavelet &
$ 4.7 $ &
$5.7$ &
$ 6.9 $ \\
\end{tabular}
\vspace{1mm}
\end{minipage} 
\hfil\hfil
\begin{minipage}{.22\textwidth}
\centering
\def\arraystretch{1.6}
 \begin{tabular}{c?ccc}
   $S$ & 
$ -3 $ & 
$ 0 $ & 
$ 3 $ \\
\Xhline{2\arrayrulewidth}
\emph{U-net} &
$ 7 $ &
$8.6$ &
$ 10.3$ \\ \hline
f-x dec. &
$ 7 $ &
$8.7$ &
$ 10.3 $ \\
\end{tabular}
\vspace{1mm}
\end{minipage}   
}
\end{table}

These results pave the way for one potential application of our method in realistic situations.
Indeed, as previously stated, having clean gathers available for the training phase is not the case when dealing with real data. Furthermore, denoising field acquisitions often require complex and computationally expensive algorithms. 

In order to overcome these issues, we recommend our strategy as a viable alternative to many standard denoising algorithms. Specifically, when a large field dataset is available, the following \red{algorithm} can be applied:
\begin{enumerate}
\item{randomly select a subset of the acquired shot gathers;}
\item{perform an accurate and computationally expensive denoising on the selected shot gathers;}
\item{train the \emph{U-net} on the selected pairs of acquired/denoised gathers;}
\item{make use of the trained \emph{U-net} to denoise the remaining data.}
\end{enumerate}
After a certain dimension of the dataset, due to the fixed computational cost for denoising the selected subset and training the \emph{U-net}, the application of the \emph{U-net}-based denoising becomes computationally cheaper than denoising the whole dataset with the standard noise attenuation algorithm. Indeed, the computational advantage of \emph{U-net} increases with the dimension of the dataset and the complexity of the denoising algorithm.

\subsection{Complete problem: Joint interpolation and denoising}
\label{subsec:interp_denoising}

 The last situation we propose is the more realistic case of study, implying additive noise corruption jointly with missing traces. We investigate two cases of study: the former exploits the same synthetic dataset of all the previous experiments, while the latter uses a different dataset, comparing our method with state-of-the-art techniques. 

\subsubsection{AWGN and uniform missing traces}
\label{subsec:uniform_awgn}

 In order to investigate if the proposed method is able to retrieve the original synthetic data, we consider the presence of AWGN and uniformly distributed missing traces.
Likewise previously done, we add noise leading into $ S \in \{3, 0, -3 \}\textrm{dB} $ and delete a percentage $ H \in \{10, 30, 50 \} $ of the available data traces for simulating \red{ seismic acquisition with irregular receiver sampling and missing traces}. This way, we generate $9$ different datasets, corresponding to various combinations of additive noise and missing traces. 
Table \ref{tbl:interp_denoising} resumes the average results obtained on shot gathers belonging to the evaluation set $ \D_E $, considering all possible combinations of missing traces and additive noise variances. 

\begin{table}[t]
	\caption{Average $\mathrm{S/N}$ [dB] achieved on gathers belonging to $\D_E$, for each dataset extracted from BP-2004 \cite{billette20052004}.}
	\label{tbl:interp_denoising}
	\centering
	\def\arraystretch{1.6}
	\resizebox{0.48\columnwidth}{!}{
		\begin{tabular}{c?ccc}
			\diagbox[width=1.5cm, height=.5cm]{\raisebox{0.5pt}{$H$} }{\raisebox{5pt}{$S$}}
			&$-3 $     &  $ 0$      & $3 $  	 	\\  \Xhline{2\arrayrulewidth}
			$ 10 $ & $12.2$ & $13.8 $ & $15.6$ \\ \hline
			$ 30$ & $11.5$ & $12.9$ & $14.4$ \\
			\hline
			$ 50$ & $10.4$ & $11.6$ & $12.9$
		\end{tabular}}
	\end{table}

\subsubsection{Comparison with a \red{recent data-driven method}}
\label{subsubsec:sota}	

 To compare our strategy with \red{a recent learning-based algorithm}, we consider the Double-Sparsity Dictionary Learning method proposed by \cite{zhu2017joint} and one strategy based on fixed dictionary transform used as baseline in \cite{zhu2017joint}, i.e., the Curvelet method.
In order to perform a fair comparison, we reproduce exactly the same synthetic example provided in \cite{zhu2017joint}.

\begin{figure*}[t]
	\centering	
	\includegraphics[width=0.75\textwidth]{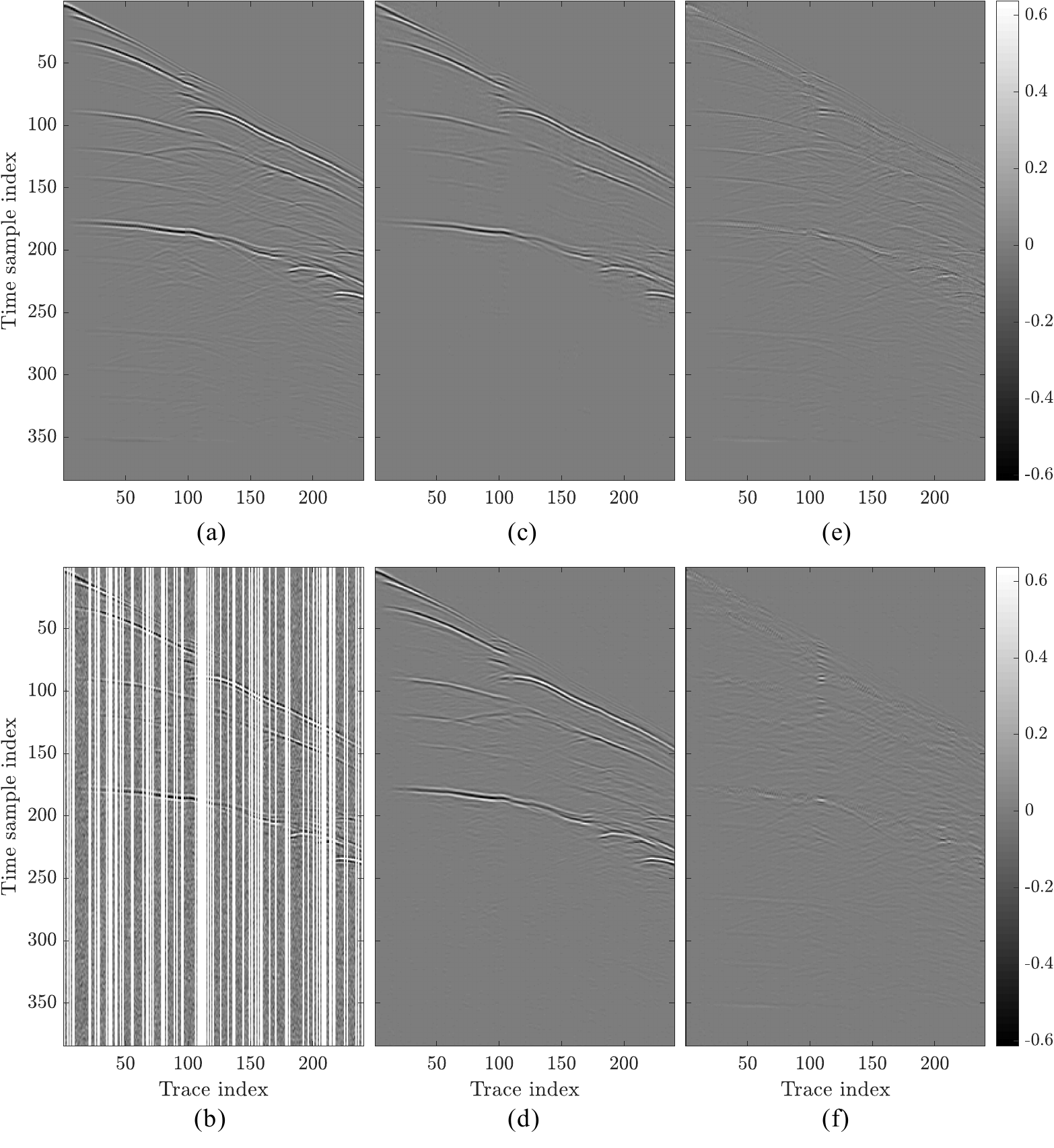}
	\caption{Shot gather with $33 \%$ of missing traces and $\sigma=0.1$. (a) shows the original gather; (b) depicts the corrupted version; (c) shows the recovered gather obtained with double-sparsity dictionary learning, $\mathrm{PS/N} = 32.1$ dB; (d) shows the recovered gather obtained with \emph{U-net}, $\mathrm{PS/N} = 33.7$ dB; (e) illustrates the reconstruction error of double-sparsity dictionary learning, i.e., (c) - (a); (f) depicts the reconstruction error of our method, i.e., (d) - (a). 
	}
	\label{fig:sota_resultsv3}
\end{figure*}
\begin{figure}[!]
	\centering	
	\includegraphics[width=0.7\columnwidth]{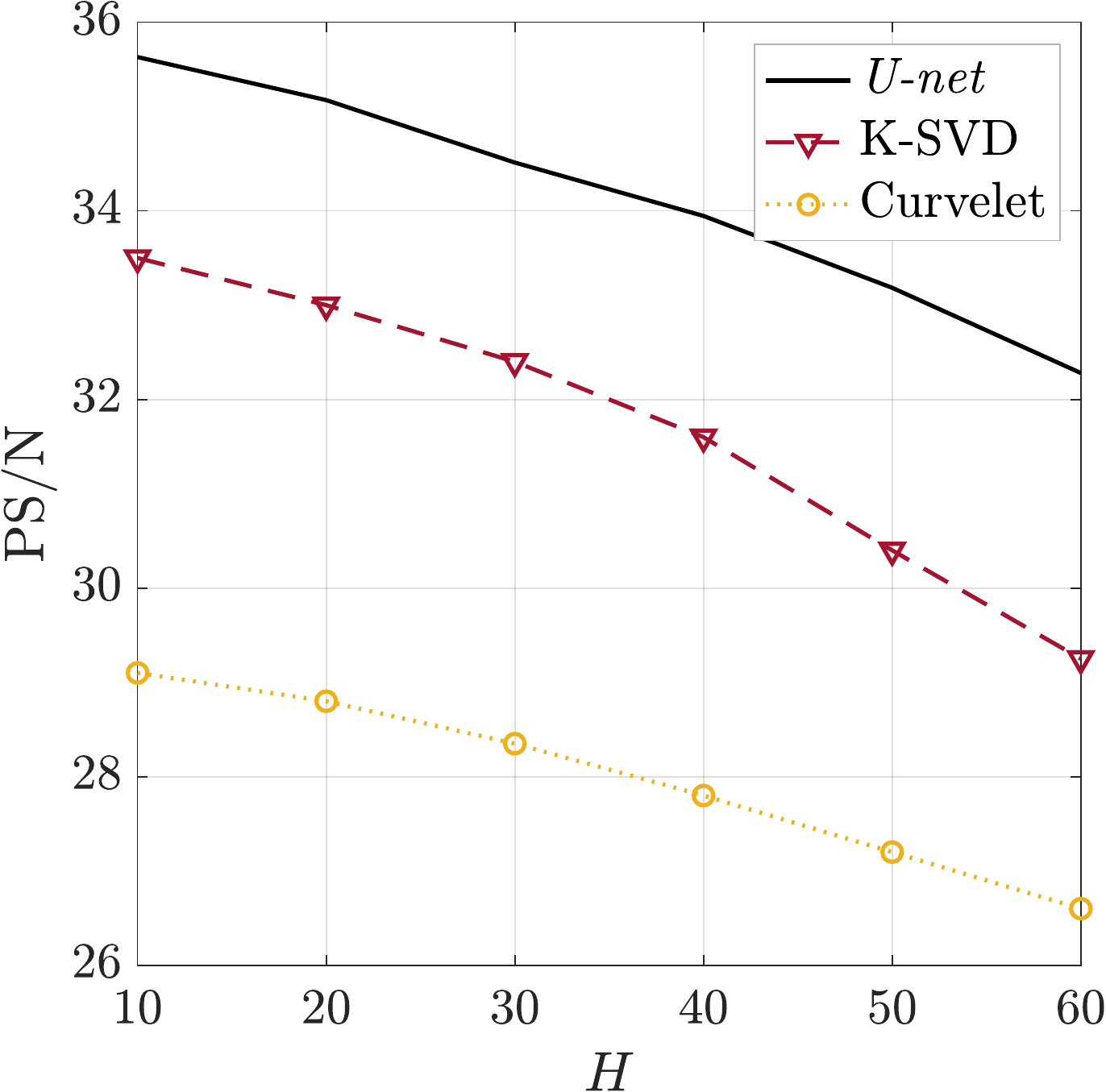}
	\caption{Results of different reconstruction methods by varying the missing traces ratio $H$ and for $\sigma=0.1$.  
	}
	\label{fig:zhu-sara_vs_H}
\end{figure}
\begin{figure}[!]
	\centering	
	\includegraphics[width=0.7\columnwidth]{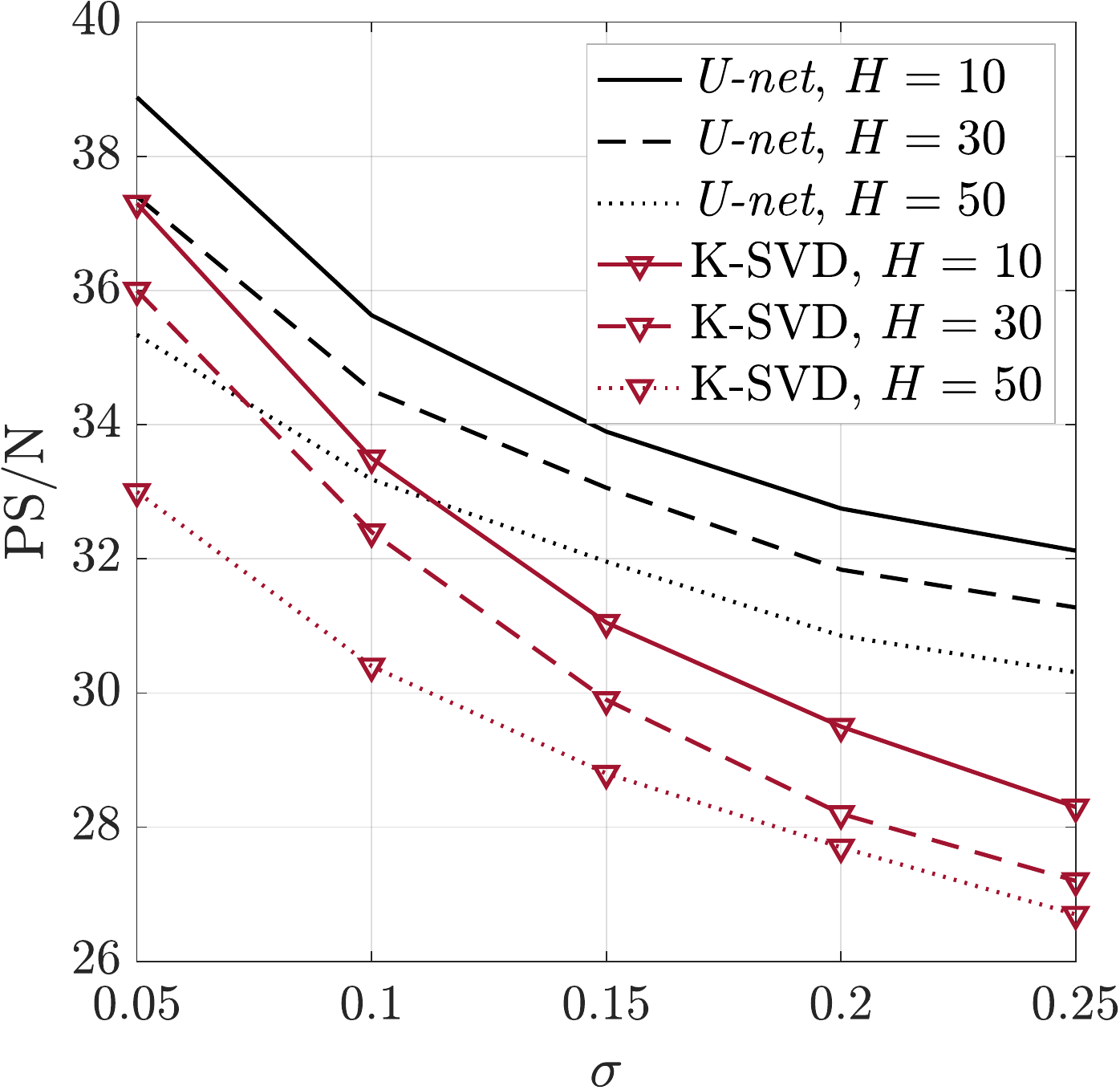}
	\caption{Results of \emph{U-net} and double-sparsity dictionary learning reconstruction methods by varying the missing traces ratio $H = \{ 10, 30, 50 \}$ and for $\sigma = 0.05:0.05:0.25$.
	}
	\label{fig:zhu-sara_vs_std}
\end{figure}

Specifically, the dataset is extracted from the BP-1997 benchmark \cite{etgen1998strike} and includes $385$ shot gathers, \red{considering the last} $240$ receivers (taking the source as reference) with $384$ samples/trace.
We add noise and \red{delete some traces} in the dataset following the procedure described therein. 
In the first stage, as done in \cite{zhu2017joint}, we normalize the range of each trace to $1$.
Then, white gaussian noise is generated, low pass-filtered with a cut-off frequency of $30\textrm{Hz}$ and finally added to the traces. 
We perform the same \red{numerical experiments} proposed in \cite{zhu2017joint}, testing plenty of noise standard deviations $\sigma \in \{0.05, 0.10, 0.15, 0.20, 0.25 \}$ and missing traces' percentages $ H \in \{10, 20, 30, 33, 40, 50, 60 \}$. 

For what concerns the training phase, we randomly select $ 250 $ shot gathers for training and validation (split in $75\%-25\%$).
In test phase, we \red{process} exactly the same image of that used in \cite{zhu2017joint}, namely the first shot gather in the dataset. Note that, in order to be fair, we never \red{use} this shot gather in the training phase.

During training, we extract from each image $153$ overlapping patches with size $128 \times 128$ and stride $(16, 14)$ along rows and columns, respectively.
This operation has been done in order to achieve training and validation sets with similar size (concerning the number of patches) to the previously shown situations.
Notice that we consider exactly the same training-validation-evaluation \red{procedure} of that depicted in Sections \ref{subsec:training} and \ref{subsec:unet_implementation}.

For comparing the results, we use the evaluation metrics proposed in \cite{zhu2017joint}, namely the peak signal to noise ratio (\red{$\mathrm{PS/N}$}), defined as
\begin{equation}
\red{\mathrm{PS/N}} = 10 \, \log_{10} \frac{s_{\max}}{\sigma^2( \I - \Ihat)},
\label{eq:psnr_sota}
\end{equation}
being $s_{\max}$ the dynamic range of the clean signal, thus fixed to $1$.

Fig.~\ref{fig:sota_resultsv3}(a) shows the original shot gather without noise added and missing traces (i.e., the ground truth of the experiment). The corrupted version of the gather with $33\% $ of missing traces and $\sigma=0.1$ is shown in Fig.~\ref{fig:sota_resultsv3}(b).
Figs.~\ref{fig:sota_resultsv3}(c)-(d) show the recovered gathers obtained with double-sparsity dictionary learning and \emph{U-net}, respectively. 
We can notice that there are some events which are well reconstructed by the \emph{U-net} while are missing in the retrieved shot gather via double-sparsity dictionary learning. 
Specifically, Figs~\ref{fig:sota_resultsv3}(e)-(f) show the error panels (i.e., the difference between the recovered images (c) and (d) and the ground truth (a)) for the results obtained with the state-of-the-art technique and \emph{U-net} respectively. 
It is quite evident that the error corresponding to double-sparsity dictionary learning is more affected by residual coherent events, meaning that those are not correctly recovered. These qualitative considerations are confirmed by the corresponding \red{$\mathrm{PS/N}$} values: $32.1$ dB for double-sparsity dictionary learning and $33.7$ dB for \emph{U-net}.
\red{Clearly, for both methods the reconstruction error is not optimum and there is still room for improvement. Nonetheless, it is worth noting that the limited size of the shot gathers (only $240 \times 384$, versus the $1152 \times 1920$ of BP-2004 benchmark and the $128 \times 1408$ of Mobil Avo Viking Graeben Line 12 dataset) could potentially undermine the ability of \emph{U-net} to learn how to describe the complex features of the clean data without modeling noise and missing data. Indeed, to be trained, CNNs usually need a substantial amount of data for achieving acceptable performances. Whether more data samples were accessible, we expect the performances to improve accordingly.} 

Fig.~ \ref{fig:zhu-sara_vs_H} displays the performances of different reconstruction methods by varying the missing traces ratio and selecting $\sigma=0.1$.   
In particular, we compare results reported in \cite{zhu2017joint} with our results, averaged over $100$ different realizations of the column pattern used for randomly deleting the traces. 
It is noticeable that we significantly outperform both the dictionary learning-based method and the Curvelet-based, gaining an average of $2.4$ dB with respect to the former strategy and $6.1$ dB to the latter one.

Fig.~\ref{fig:zhu-sara_vs_std} reports the achieved results for a plurality of noise standard deviations.
The performances of the \emph{U-net} are significantly superior than those of dictionary learning-based strategy, in all the examined cases.
Moreover, our method reveals to be more robust in presence of strong noise. 
As a matter of fact, as noise standard deviation $\sigma$ increases, the curves related to state-of-the-art method decay in a worse fashion than ours, to the point that we can achieve $\red{\mathrm{PS/N}} = 30.3$ dB for $H = 50$ and $ \sigma = 0.25$, against the $26.7$ dB of the dictionary learning-based technique.

\section{Conclusions}
\label{sec:conclusions}

\hspace{\parindent} We proposed a method for reconstruction of corrupted seismic data, focusing on noise attenuation and interpolation of missing pre-stack data traces in the shot-gather domain. 
In particular, we considered random noise \red{cases} with different statistics and a \red{variety} of missing traces distributions. 
Our approach 
makes use of a convolutional neural network architecture for interpolation and denoising of 2D shot gathers, showing performance improvements, \red{in terms of S/N ratio}, with respect to recent solutions for joint denoising and interpolation.

Results achieved on controlled synthetic experiments demonstrate that the proposed method is a promising strategy for seismic data pre-processing. 
\red{The method is capable of effectively and efficiently restore 2D corrupted data, and it is also able to deal with the task of spatially upsampling the shot gathers.}
Moreover, once the network training procedure is completed, processing data with our strategy is also quite efficient in terms of computational effort.

We examined also the potential application of our methodology on field data for production environments. 
In this situation, it is interesting to notice that the proposed algorithm can be used also to emulate the effect of more time consuming classical data pre-processing strategies.



Future work will be devoted to investigating issues related to denoising of field data, exploring for instance the feasibility of a transfer learning procedure by training convolutional neural networks on properly designed synthetic data and testing on field data. 
Moreover, investigations are needed for denoising of more challenging types of coherent noise and artifacts affecting seismic data (e.g., ground roll in land data) \red{and different kinds of missing data (e.g., missing short offsets and cross-line upsampling).}

Further studies on the network architectures and loss functions (\red{e.g., considering the use of variational autoencoders, generative adversarial networks, etc.}) could relax the need of corrupted/uncorrupted pairs of gathers for the training set, thus helping in dealing with the problem of building a training dataset for denoising.

\red{In addition, investigations are needed to carefully examine the effect of the proposed strategy on standard processing and imaging algorithms (e.g., surface related multiple elimination, reverse time migration, etc.).}

Finally, other aspects we would like to examine for moving towards production environments are \red{exploiting the similarity between adjacent gathers to improve the reconstruction performances} and the extension of the proposed procedure to 3D data. 

In the light of the promising results achieved with the proposed architecture of convolutional neural network, we believe this tool can pave the way for even more efficient and accurate solutions.
\section{Glossary on CNNs}
\label{sec:glossary}

Deep learning and CNNs are still relatively new topics for the geophysical community. In order to ease the readability of the paper, this glossary reports the definition of some standard CNN-related terms which have been used throughout the paper.
Since our experiments have been conducted using Keras framework, for a thorough reading, please refer to \cite{keras}. 
\begin{itemize}
    \item\textit{Activation function}: it is a scalar function (typically nonlinear) that is pointwise applied to the output of a layer.
    \item \textit{Batch}: the set of data (e.g., images or image-patches) used for one iteration (i.e., a gradient update step) during model training.
    \item \textit{Batch Normalization (BN)}: this layer applies a transformation to the data in order to maintain the mean close to 0 and the standard deviation close to 1, for each batch.
    \item \textit{Concatenate}: layer that concatenates a list of inputs, which have all the same dimensions except for the concatenation axis.
    \item \textit{Convolutional layer}: it is the core layer of CNN architecture. The convolutional layer consists of nodes that perform a convolution between an input and a set of filters (convolutional kernels). In the 2D case, the input image (or image-patch) to an arbitrary convolutional layer can be defined as $\I_{in} (x, y, c)$, with $c \in [1, ..., N_{c}]$. 
    Specifically, $x$ and $y$ are the 2D image dimensions and $c$ the so-called ``channel" dimension. 
    Each convolutional layer has a set of $N_j$ bidimensional filters: a generic filter is defined as $\w_{j}(x ,y, c)$. 
    The output of the layer will be:
    \begin{multline}
    \I_{out} (x ,y, j)= J \bigg( \mathbf{B}_{j}(x, y) +  \\ \sum_{c=1}^{N_c}\I_{in}(x,y,c) \ast  \w_{j}(x ,y, c) \bigg)   \quad \forall j \in [1, ..., N_j],
    \end{multline}
    where $\mathbf{B}_{j}(x,y)$ is a bias matrix, $\ast$ is the symbol for the 2D convolution on $x$ and $y$ dimensions, and $J(\cdot)$ is an activation function. 
    The filter weights are set during the learning stage. Notice that the input size $N_c$ depends on the number of filters on the previous convolutional layer.
    \item \textit{Dropout}: this layer randomly sets to zero the update of some network node at each iteration during the training phase. This can help prevent overfitting. In our experiments, Dropout layer sets to zero the $50\%$ of neurons. 
    \item \textit{Epoch}: one pass, during network training,  over the entire training dataset. Since the CNN model updates every time a single batch of data is processed, the model is updated multiple times (i.e., the number of batches) during one epoch.
    \item \textit{Leaky ReLU}: it is a leaky version of a Rectified Linear Unit. Leaky ReLU is used to avoid too many zeros flooding the network; each negative signal sample $x$ is converted to $\alpha \cdot x $, with $\alpha$ typically small (e.g., $\alpha = 0.2$ in our experiments):
    $$
        \textrm{LeakyReLU}(x) = \left\{\begin{matrix}
        x &\textrm{if} \ x>  0 \\ 
         \alpha \cdot x &\textrm{if} \ x\leq  0.
        \end{matrix}\right.
    $$
    \item \textit{Learning rate}: this parameter controls how much to adjust the CNN weights with respect to the loss gradient. The lower the learning rate, the slower we descend towards the minimum of loss function. The corresponding concept in seismic inverse problem terminology is that of update step-size.
    \item \textit{Loss}: it is the cost function which measures how far the network predictions are from the desired results. This is the function to be minimized during the network training.
    \item \textit{Patience}: the number of epochs with no improvement in the validation loss function (or validation accuracy metrics) after which the training will be stopped.
    \item \textit{ReLU}:  the Rectified Linear Unit is an activation function which converts to $0$ the negative signal samples, leaving untouched the positive ones:
    $$
    \textrm{ReLU}(x)=\max\left(0,x\right).
    $$
    \item \textit{Stride}:
    in a convolutional layer, applying a stride $k$ in the $i^{th}$ dimension means that convolutions are computed by moving the kernel with a step of $k$ samples along the $i^{th}$ dimension. The output of a convolution with stride $k$ in the $i^{th}$ dimension is equivalent to downsampling at rate $k$ along the $i^{th}$ dimension of the output of a standard convolution.
\end{itemize}

\ifCLASSOPTIONcaptionsoff
  \newpage
\fi

\bibliographystyle{IEEEtran}
\bibliography{bibliography}
\end{document}